%% file: main.tex
\documentclass[10pt,twocolumn,letterpaper]{article}

\usepackage[pagenumbers]{wacv} 

\input{preamble}

\definecolor{wacvblue}{rgb}{0.21,0.49,0.74}
\usepackage[pagebackref,breaklinks,colorlinks,allcolors=wacvblue]{hyperref}

\def\wacvPaperID{938} 
\def\confName{WACV}
\def\confYear{2027}

\title{Rectifying Mask via Entropy for Distractor-Free 3DGS in Ambiguous Scenarios}

\author{
    Wongi Park$^{1}$ \quad
    Jiyeon Lim$^{2}$ \quad
    Minjae Lee$^{3}$ \quad
    Myeongseok Nam$^{4}$ \quad
    Seongjun Choi$^{5}$ \quad \\
    Jungwoo Kim$^{6}$ \quad
    Soomok Lee\textsuperscript{\dag}$^{7}$ \quad 
    William J. Beksi\textsuperscript{\dag}$^{8}$ \quad
    Sang-Hyun Lee\textsuperscript{\dag}$^{1}$ \quad
    \vspace{2mm}
\\
 \normalsize
 \textsuperscript{1}Ajou Univerity \;\;
 \textsuperscript{2}Samsung Electronics \;\;
 \textsuperscript{3}Georgia Institute of Technology \;\; 
 \textsuperscript{4}GenGenAI \;\; \\
 \normalsize
 \textsuperscript{5}Yonsei University \;\;
 \textsuperscript{6}Seoul National University \;\;
 \textsuperscript{7}Kennesaw State University 
 \quad
 \textsuperscript{8}University of Texas at Arlington 
\\
   {\github \href{https://github.com/WongiPark0628/RefineSplat}{{\text{Code}}}}
\quad
{\worldwideweb \href{https://wongipark0628.github.io/RefineSplat/}{{\text{Project page}}}}
\quad
{\huggingface \href{https://huggingface.co/datasets/kalelpark/RefineSplat}{{\text{Dataset}}}}
}

\begin{document}

\input{section/abstract}

\input{section/intro}
\input{section/related}
\input{section/preliminary}
\input{section/methods}

\input{section/experiments}
\input{section/ablation}
\input{section/conclusion}

\clearpage
\input{appendix}
\clearpage
{
    \small
    \bibliographystyle{ieeenat_fullname}
    \bibliography{main}
}

\end{document}

%% file: preamble.tex
\usepackage{graphicx}
\usepackage{booktabs}

\usepackage[accsupp]{axessibility}  
\usepackage{amsmath}
\usepackage{amssymb}
\usepackage{multirow}
\usepackage{caption}
\usepackage{algpseudocode}
\usepackage[noEnd=false]{algpseudocodex}
\usepackage{subcaption}
\usepackage[table]{xcolor}
\usepackage{multirow}
\usepackage{amsmath}
\usepackage{setspace} 
\usepackage{lipsum}
\usepackage{kotex} 
\usepackage[dvipsnames]{xcolor}
\usepackage{amssymb}
\usepackage{pdfrender}
\usepackage{bbding}
\usepackage{pifont}
\usepackage{amsfonts}
\usepackage{bbm}
\usepackage{graphicx}
\usepackage{wrapfig}
\definecolor{cgood}{HTML}{90C060}
\usepackage[ruled,linesnumbered]{algorithm2e}

\newcommand*{\boldcheckmark}{%
  \textpdfrender{
    TextRenderingMode=FillStroke,
    LineWidth=.5pt, %
  }{\checkmark}%
}

\DeclareSymbolFont{extraup}{U}{zavm}{m}{n}
\DeclareMathSymbol{\varheart}{\mathalpha}{extraup}{86}
\DeclareMathSymbol{\vardiamond}{\mathalpha}{extraup}{87}

\newcommand{\bgcolor}[2]{\setlength{\fboxsep}{0pt}\colorbox{#1}{\strut #2}}
\newcommand{\BGcolor}[3][HTML]{\definecolor{mycolor}{HTML}{#2}\bgcolor{mycolor}{#3}}

\newcommand{\metrictablefirst}[1]{{\BGcolor{87faba}{#1}}}
\newcommand{\metrictablesecond}[1]{{\BGcolor{bafaab}{#1}}}
\newcommand{\metrictablethird}[1]{{\BGcolor{e9faab}{#1}}}

\newcommand{\worldwideweb}{\raisebox{-1.5pt}{\includegraphics[height=1.05em]{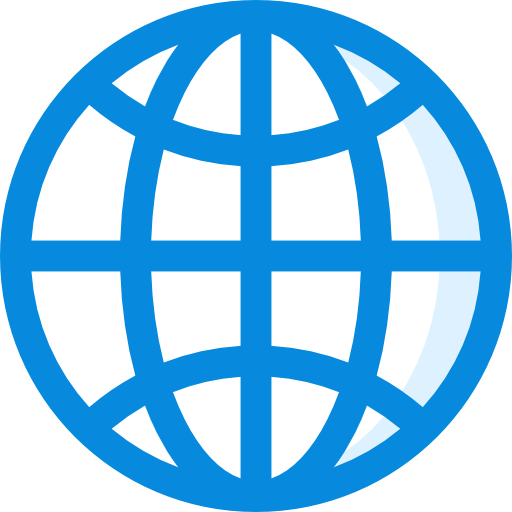}}\xspace}
\newcommand{\github}{\raisebox{-1.5pt}{\includegraphics[height=1.05em]{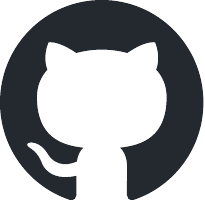}}\xspace}
\newcommand{\huggingface}{\raisebox{-1.5pt}{\includegraphics[height=1.05em]{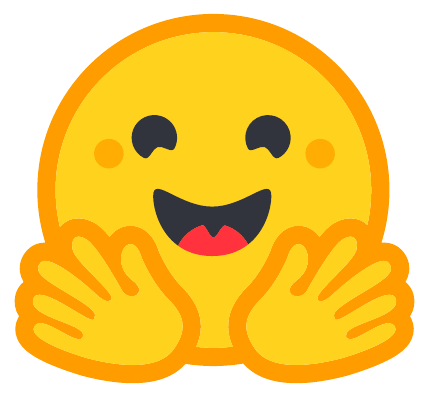}}\xspace}

%% file: section/abstract.tex
\input{tab_fig/fig_main}

\renewcommand{\thefootnote}{}
\footnotetext{\textsuperscript{\dag}Corresponding authors}

\begin{abstract} 
We present RefineSplat, a systematic framework that effectively constructs transient masks to identify diverse ambiguous distractors. To do this, we qualitatively and quantitatively analyze issues and propose a novel entropy-aware adaptive masking method. Unlike existing approaches that struggle to distinguish transient elements from static scenes due to color or semantic ambiguity, RefineSplat captures ambiguous distractors leveraging entropy and instance masks. Furthermore, we propose a simple yet effective entropy-aware density control to align Gaussians in ambiguous scenarios considering Entropy-aware positional gradients. Additionally, to rigorously validate our method, we first create and release the Ambiguous wild dataset, including 18 scenes where distractors and static scenes are hard to distinguish due to color or semantic resemblances. Experimental results on various datasets demonstrate that RefineSplat shows state-of-the-art performance, showing distractor-free novel view synthesis. 
\end{abstract}

%% file: tab_fig/fig_main.tex
\twocolumn[{%
    \renewcommand\twocolumn[1][]{#1}%
    \maketitle
    \begin{center}
        \centering
        \captionsetup{type=figure}        \includegraphics[width=\textwidth]{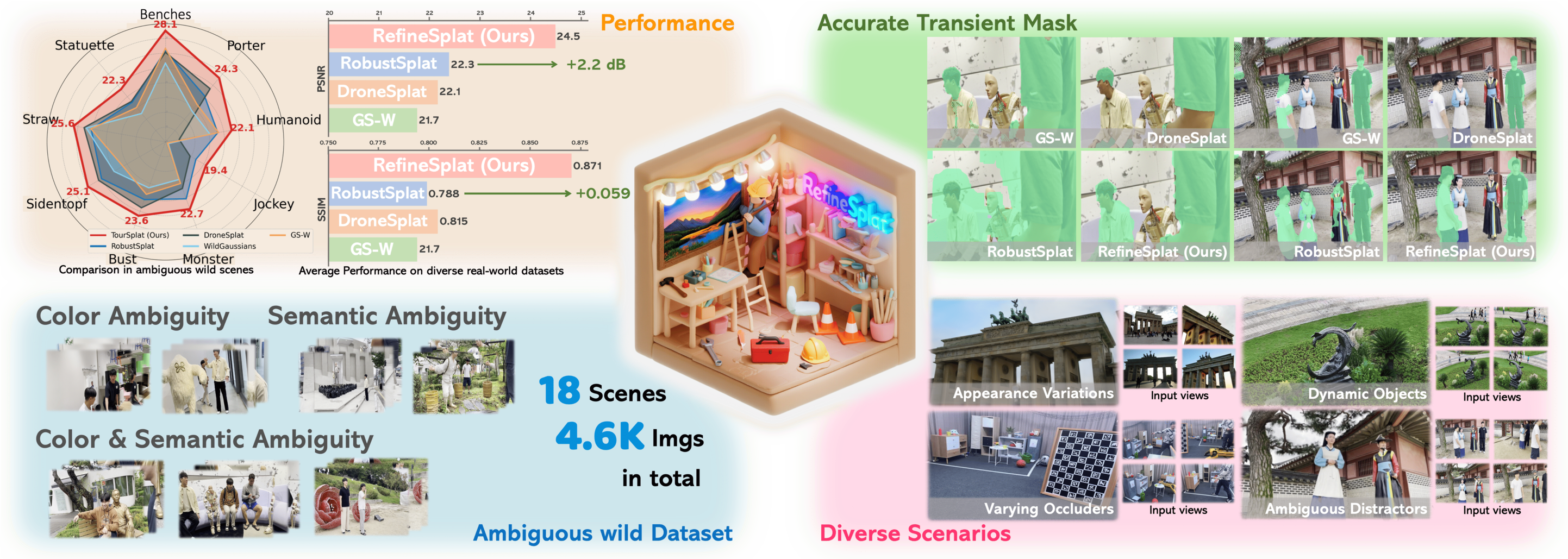}
        \vspace{-2.5em}
        \caption{Given real-world environments, RefineSplat effectively renders 3D novel view synthesis without ambiguous distractors. Our method effectively handles several challenging distractors (e.g., appearance variations, varying occlusions, dynamic objects, ambiguous distractors). To the best of our knowledge, this is the first work to pave the way for novel view synthesis in ambiguous real-world scenarios.} 
    \end{center}
}]


%% file: section/intro.tex
\section{Introduction}
Realistically rendering 3D scenes with 3D Gaussian Splatting (3DGS) \cite{kerbl20233d} and Neural radiance field (NeRF) \cite{mildenhall2021nerf} from diverse real-world scenarios is a challenging and fundamental
problem \cite{martin2021nerf, zhang2024gaussian, kulhanek2024wildgaussians, park2026forestsplats, fu2025robustsplat}. Unlike conventional methods \cite{yu2024mip, hu2023tri, cheng2024gaussianpro} that primarily focus on constrained images which do not include distractors, this problem is characterized by the highly unstructured nature of the inputs (\eg, appearance variations, diverse distractors, dynamic objects, etc). To address these issues, existing methods \cite{sabour2024spotlesssplats, ren2024nerf, tang2025dronesplat} have tried to tackle unconstrained images, broadly falling into three paradigms. (1) Residual-based masking \cite{li2023nerf, martin2021nerf, zhang2024gaussian, sabour2023robustnerf} relies on photometric error to generate transient masks. (2) Semantic-based masking \cite{zheng2025wildgs, kulhanek2024wildgaussians, fu2025robustsplat} exploits extracted features from DINOv2 \cite{oquab2023dinov2} to identify distractors at the semantic level. (3) Heuristic-based masking \cite{tang2025dronesplat, chen2024nerf, markin2024t} leverages extracted masks from SAM \cite{kirillov2023segment} using photometric error as a condition.

\input{tab_fig/fig_motivation}

\vspace{0.2\baselineskip}
\noindent \textbf{Challenges.} Although prior methods \cite{wang2024desplat, lin2024hybridgs, kulhanek2024wildgaussians, tang2025dronesplat} show impressive results across real-world scenarios, these methods are still impeded by three limitations. (1) Leveraging photometric error makes it difficult to produce precise transient masks due to ambiguities, especially in cases where the background and distractors share similar colors. (2) Positional gradients based on photometric error for density control also degrade rendering quality by misaligning Gaussians in ambiguous scenarios. (3) Semantic-based masking \cite{kulhanek2024wildgaussians, fu2025robustsplat, ren2024nerf} , only leveraging extracted features from DINO \cite{oquab2023dinov2}, struggles with the challenge of semantic level similarity (e.g., moving vs stationary vehicles, a person vs a human-like statue).

\vspace{0.2\baselineskip}
\noindent \textbf{Motivations and analysis.} To understand how this issue emerges, we visualize our motivation in Fig.~\ref{fig:motivation} and Fig.~\hyperref[fig:transient_mask]{\ref{fig:transient_mask}-(a)}. Through this observation, we indicate that transient masks fail to capture distractors due to the color or semantic ambiguity between transient and static elements. Furthermore, we quantitatively and qualitatively confirm that the density control based on photometric error often misaligns Gaussians, as shown in Eq.~\ref{fig:eq_problem} and Fig. \ref{fig:entropy-aware-densify}. These limitations misalign Gaussians, degrading rendering quality.

\vspace{0.2\baselineskip}
\noindent \textbf{Proposed solution.} To address these limitations, we introduce RefineSplat, a novel framework that leverages entropy and boundary segments to capture ambiguous distractors. Specifically, we propose the entropy-aware adaptive masking, which utilizes entropy and instance masks. Furthermore, we propose the simple and yet novel entropy-aware density control strategy to mitigate misaligned Gaussians and align Gaussians, reducing artifact issues. To rigorously validate our method, we create and release the Ambiguous wild dataset, including 18 scenes where distractors and static scenes are difficult to identify due to color or semantic ambiguity. Through extensive experiments, we demonstrate that RefineSplat outperforms existing methods. 

\vspace{0.2\baselineskip}
\noindent \textbf{Key distinctions.} To the best of our knowledge, RefineSplat is the first approach to capture ambiguous distractors leveraging entropy in real-world scenarios. This is not a trivial extension of prior works, as it (1) formulates entropy-aware methods to capture ambiguous distractors; (2) utilize entropy magnitude to align Gaussians; and (3) constructs the Ambiguous wild dataset and consistently show impressive results in diverse real-world scenarios, whereas prior works are limited to two scenarios. Our major contributions can be summarized as follows: \begin{itemize} \item
We introduce RefineSplat that leverages Entropy-aware adaptive masking, which integrates entropy and instance masks to capture ambiguous distractors.  \item 
We propose the Entropy-aware density control that effectively aligns Gaussians, mitigating artifact issues. \item
We create and release the Ambiguous wild dataset, comprising 18 challenging indoor and outdoor scenes where distinguishing distractors from static elements is particularly difficult due to color or semantic similarities. \item
Extensive results demonstrate that RefineSplat achieves superior results over the state-of-the-art methods across diverse real-world scenarios.
\end{itemize}



%% file: tab_fig/fig_motivation.tex
\begin{figure}[!t]
    \centering
    \includegraphics[width=\linewidth]{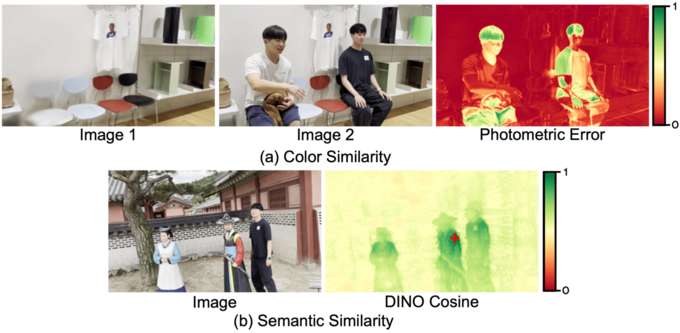}
    \vspace{-2em}
    \caption{Motivation. We color the normalized values, ranging from \BGcolor{5fbaa8}{o}\BGcolor{88cfa4}{n}\BGcolor{b2e0a2}{e}\BGcolor{d7ef9b}{ }\BGcolor{eff8a6}{t}\BGcolor{ffffbf}{o}\BGcolor{feeb9e}{ }\BGcolor{fdd380}{z}\BGcolor{fdb466}{e}\BGcolor{f88d51}{r}\BGcolor{f06744}{o}. Ambiguous transient elements are difficult to distinguish due to color or semantic similarity.} 
    \label{fig:motivation}
    \vspace{-1.5em}
\end{figure}

%% file: section/related.tex
\section{Related Work}
\noindent \textbf{Novel view synthesis for unconstrained scenes.} Most of the primary novel view synthesis methods \cite{mildenhall2021nerf, kerbl20233d, shin2025locality, li2025scenesplat, gao2025surfacesplat} rely on the assumption of transient-free environments with consistent scenes. However, diverse real-world scenarios often violate these assumptions due to distractors. Numerous existing methods \cite{martin2021nerf, mildenhall2021nerf,yang2023cross, li2023nerf} try to tackle these issues by constructing transient masks. NeRF-W \cite{martin2021nerf} exploits an uncertainty embedding per view and a transient field which represent distractors to construct transient masks. CR-NeRF \cite{yang2023cross} propose a cross-ray paradigm and a grid sampling strategy for efficient appearance modeling. NeRF-MS \cite{li2023nerf} also utilizes transient embeddings and multi-sequence images captured with triplet loss. While NeRF-based methods \cite{martin2021nerf, mildenhall2021nerf,yang2023cross} make effective transient masks, they still limit real-time rendering. 3DGS-based methods \cite{kerbl20233d, yang2024depth, xu2024wild, kulhanek2024wildgaussians}, which is primarily advent, explicitly represent the 3D scene with differentiable rasterization, showing real-time rendering. Wild-GS \cite{xu2024wild} utilizes extracted depth maps from frozen Depth anything \cite{yang2024depth} to capture transient elements and preserve geometric consistency. WildGaussians \cite{kulhanek2024wildgaussians} also utilizes extracted semantic features from frozen DINOv2 \cite{oquab2023dinov2} to tackle transient elements. SpotlessSplats \cite{sabour2024spotlesssplats} leverage the power of diffusion model with clustering process to identify transient elements using photometric errors. Several methods \cite{wang2025desplat, lin2024hybridgs, park2026forestsplats} represent both static and transient elements by constructing transient fields. Specifically, ForestSplats  \cite{park2026forestsplats} shows efficient memory usage to capture transient elements by adapting deformable MLPs. However, existing approaches \cite{li2023nerf, ren2024nerf, tang2025dronesplat, kulhanek2024wildgaussians} primarily focus on either semantic features or residual maps, which makes it difficult to capture ambiguous distractors. In contrast, our method identifies ambiguous distractors by leveraging entropy and instance levels in real-world scenarios.

\vspace{0.2\baselineskip}
\noindent \textbf{Adaptive density control.} Unlike NeRF \cite{mildenhall2021nerf} with fixed parameter counts, 3DGS \cite{kerbl20233d} represents 3D scenes using Gaussian primitives from 3D points. Recently, several works improve the densification strategy for various purposes: compressing \cite{lee2024compact, lee2025optimized, fan2023lightgaussian, zhang2025gaussianspa}, structural encoding \cite{chen2024hac, lu2024scaffold, zhang2025neural}, and controlling density \cite{bulo2024revising, zeng2025frequency, wang2025steepest, ye2024absgs}. However, most methods aim to work in constrained images, but their application to unconstrained images remains challenging due to transient elements. To tackle these challenges, several methods \cite{park2026forestsplats, jiang2024robust} have emerged, offering robust densification strategies, reducing redundant Gaussians. 
RobustSplat \cite{jiang2024robust} mitigates redundant Gaussians by leveraging delayed Gaussian growth. Unlike prior methods, we introduce an entropy-aware density control that aligns Gaussians, mitigating artifact issues in real-world scenarios.

\input{tab_fig/fig_motivate}

%% file: tab_fig/fig_motivate.tex
\begin{figure}[!t]
    \centering
    \includegraphics[width=\linewidth]{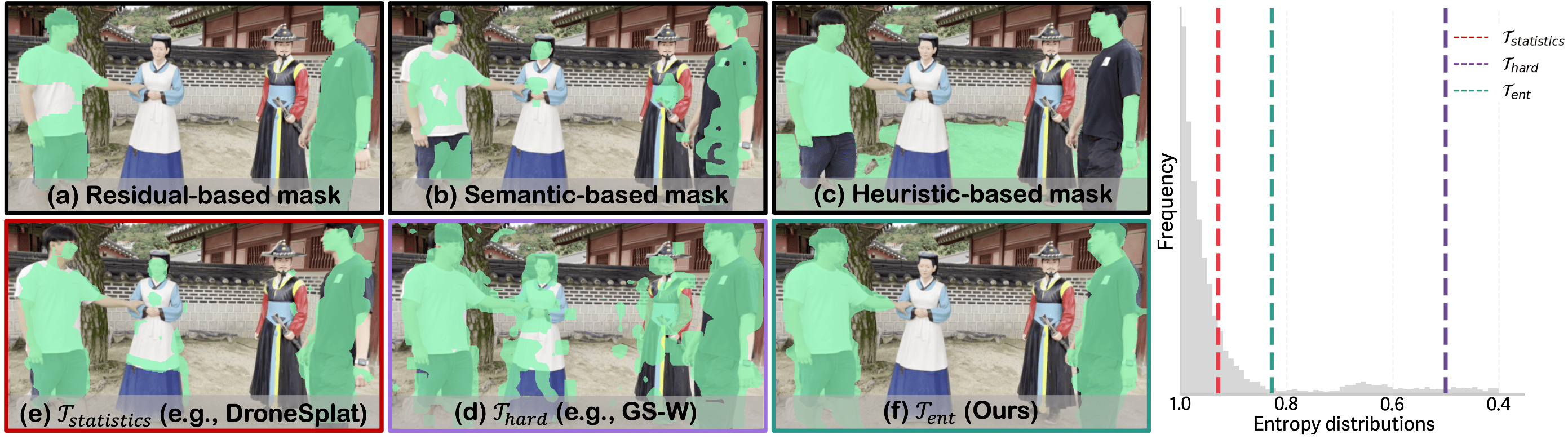}
    \vspace{-2em}
    \caption{Effectiveness of transient masks and thresholds. (a) and
    (b) fail to capture distractors with similar color and semantics to static elements. (c) only focuses on extracted invariant masks from
    SAM before training. For the thresholds, (d) overestimates distractors. (e) is hard to capture ambiguous distractors, as it is heavily dependent on averages. However, (f) precisely captures ambiguous distractors in a skew distribution. (check color boundary)} 
    \label{fig:method_motivation}
    \vspace{-1.5em}
\end{figure}

%% file: section/preliminary.tex
\section{Preliminary}
\textbf{3D Gaussian Splatting.} 3D Gaussian Splatting (3DGS) represents a 3D scenes as an explicit set of anisotropic 3D Gaussians $\{ \mathcal{G}_i \}$, each having a position $ \mu_{i} \in \mathbb{R}^{3}$, covariance matrix $ \Sigma_{i} \in \mathbb{R}^{3 \times 3}$, which is decomposed into scaling matrix $S \in \mathbb{R}^{3}$ and rotation matrix $R \in \text{SO(3)}$, opacity $\alpha_{i} \in [0, 1]$, and view-dependent colors $c_{i} \in \mathcal{C}^{N_{\text{SH}}}$ represented via spherical harmonics (SH) coefficients $N_{\text{SH}}$. To conduct 3D novel view synthesis, 3D Gaussians $\{ \mathcal{G}_i \}$ are mapped to the image plane through the viewing transformation $W \in \mathbb{R}^{3 \times 3}$ and covariance matrix $\Sigma'_{i} = JW \Sigma_{i} W^{T} J^{T} $ where $J \in \mathbb{R}^{2 \times 3}$ is the Jacobian of the Taylor approximation of the projective transformation. The pixel color \(C(\boldsymbol{p})\) is computed by sorting the Gaussians according to depth and alpha blending:
\begin{equation}
  C(\boldsymbol{p}) =  \sum_{i=1}^N \,c_i\,\alpha_i\,\prod_{j=1}^{i-1}\,\bigl(1 - \alpha_j\bigr)
  ,
\end{equation} where $\alpha_i = \sigma(o_i)\exp(-\frac{1}{2}(p_{i} - \mu'_{i})^\top\Sigma'^{-1}_{i}(p_{i} - \mu'_{i}))$. $p_{i}$ and $\mu_{i}'$ are a pixel coordinate and a position of 2D Gaussian. All attributes of Gaussian $\{ \mathcal{G}_i \}$ are optimized by minimizing the loss $\mathcal{L}_{\mathrm{GS}}$ between rendered image $\hat{I} \in \mathbb{R}^{3 \times H \times W}$ and ground truth image $I^{gt}  \in \mathbb{R}^{3 \times H \times W}$ as follows: \begin{equation}
\mathcal{L}_{\mathrm{GS}}=(1-\lambda) \mathcal{L}_1(\hat{I}, I^{gt})+\lambda \mathcal{L}_{\text {D-SSIM}}(\hat{I}, I^{gt} ), \end{equation}
where $\mathcal{L}_{1}$ is an $L_{1}$ loss, $\mathcal{L}_{\text{D-SSIM}}$ is a SSIM loss and $\lambda$ is a weighting factor set to $0.2$. All 3D Gaussians are initialized using a sparse point obtained from the Structure-from-Motion (SfM) approach such as COLMAP \cite{schoenberger2016mvs}.

\vspace{0.2\baselineskip}
\noindent \textbf{Adaptive density control.} 3DGS and follow-up methods \cite{yu2024mip, bulo2024revising} rely on the Adaptive Density Control (ADC) module to densify 3D Gaussians. ADC module periodically operates to densify 3D Gaussians considering positional gradient $||\frac{\partial\mathcal{L}_{\mathrm{GS}}}{\partial\mu}||$ based on photometric error.

%% file: section/methods.tex
\input{tab_fig/fig_method}
\section{Method}
Motivated by the preceding analysis, we introduce RefineSplat, which effectively captures ambiguous distractors that are difficult to distinguish due to color or semantic resemblance. As shown in Fig. \ref{fig:methods}, our method consists of two main components: (1) the entropy-aware adaptive masking, which leverages entropy and instance masks (Sec. \ref{subsec:Entropy-awareAdaptiveMasking}). (2) Entropy-aware density control that leverage entropy magnitude to clone and splits Gaussians. Furthermore, we align Gaussians merging overlapping Gaussians (Sec. \ref{subsec:Entropy-awareDensityControl}). Moreover, we introduce the consistency regularization to rectify incorrect instance masks (Sec. \ref{subsec:Regularization}).

\subsection{Entropy-aware adaptive masking}
\label{subsec:Entropy-awareAdaptiveMasking}
\noindent \textbf{Entropy-aware local adaptive masking.} To tackle ambiguous distractors, we leverage the 2D segmentation model \cite{cheng2023tracking} to obtain instance masks $\{ \mathcal{M}^1_i, \mathcal{M}^2_i, \mathcal{M}^3_i, \cdot\cdot\cdot, \mathcal{M}^K_i \} = S(I^{gt}_{i})$, where $K$ is the number of instance masks, and $\mathcal{M}_i^j$ denotes the mask of the $j$-th object in the image. Given these instance masks, we calculate the average entropy for each instance as: 
\begin{equation} 
\bar{\mathcal{H}}^j_i = \frac{\Sigma_{p \in \mathcal{M}^j_i}(\mathcal{H}_i(p))}{|\mathcal{M}^j_i|},\;
\mathcal{H}_i(p) = - \sigma(\mathcal{E}_i(p)) \log \sigma(\mathcal{E}_i(p)),    
\end{equation} where $\mathcal{E}_i$ is calculated as $\mathcal{E}_i = \psi_{id}(D_{\theta}(\hat{I}))$, $|\mathcal{M}^j_i|$ denotes the number of pixels in the instance mask, $\sigma(\cdot)$ denotes the sigmoid function, and $p \in \mathcal{M}^j_i$ indicates that $p$ is one of the pixels contained within the mask $\mathcal{M}^j_i$. Specifically, we utilize a frozen MLPs $\psi_{id}$ to extract entropy by leveraging extracted semantic features from DINOv2 $D_{\theta}$, following prior works \cite{kulhanek2024wildgaussians, zheng2025wildgs, ren2024nerf}. Although straightforwardly utilizing $\bar{\mathcal{H}}^j_i$ with fixed threshold is seemingly beneficial, leveraging $\bar{\mathcal{H}}^j_i$ with fixed threshold has limitations in clearly identifying distractors. To address these issues, several methods \cite{ren2024nerf, sabour2024spotlesssplats, tang2025dronesplat, wang2025desplat} leverage thresholds based on statistical measures (e.g., mean, variance, percentile). However, these methods still struggle to handle the skewed distributions of $\bar{\mathcal{H}}^j_i$ that primarily arise in ambiguous scenarios where distractors have similar colors or semantic levels to static elements, as presented in Fig.  \ref{fig:method_motivation}. To overcome these problems, we introduce the adaptive entropy threshold $\mathcal{T}_{ent}$ to identify ambiguous distractors with Shannon entropy \cite{shannon1948mathematical} during training. For each iteration, we compute the optimal threshold $\mathcal{T}_{ent}$ by maximizing the sum of entropy between the lower and upper subsets of the instance entropy distribution $\bar{\mathcal{H}}^j_i$ as follows:
\begin{equation}
\mathcal{T}_{ent} = \arg \max_{t} \left( (1 - w_{2}) \mathcal{E}(\bar{\mathcal{H}}^j_{i} \geq t) + w_{2} \mathcal{E}(\bar{\mathcal{H}}^j_{i} < t) \right),
\end{equation}
where $w_2$ is a positive learnable weight factor used to adaptively adjust threshold. Finally, the local adaptive mask $\mathcal{M}_{local}$ is constructed by selecting instances that exceed this optimized threshold as follows:
\begin{equation}
\mathcal{M}_{local} = \{ \mathcal{M}^j_i \mid \bar{\mathcal{H}}^j_i > \mathcal{T}_{ent}, j \in {1, \dots, K} \}.
\end{equation}

\vspace{0.2\baselineskip}
\noindent \textbf{Entropy-aware global adaptive masking.} Although relying solely on robust masks $\mathcal{M}_{local}$ yields impressive results, this approach remains limited in handling fine-grained transient elements and incorrect extracted masks from SAM \cite{kirillov2023segment}. To address these, we leverage entropy and semantic features to globally identify subtle distractors as follows:
\begin{equation}
    \mathcal{M}_{global} = \mathbbm{1}\{ ( \frac{\mathcal{U}_i}{\mathcal{H}_i} \odot \text{B}_{3\times3}) > \mathcal{T}_{glb}  \},
\end{equation}where $\text{\small} \mathcal{U}_i = \frac{2}{(\cos(D_\theta(I_i^{gt}), D_\theta(\hat{I}_i))+ 1)}$ is an uncertainty map, $\mathcal{T}_{glb}$ is defined $\text{\small} \mathcal{T}_{glb} = \mathbb{E}\left[\frac{\mathcal{U}_i}{\mathcal{H}_i}\right]+\operatorname{Var}\left[\frac{\mathcal{U}_i}{\mathcal{H}_i}\right]$, and  $\text{\small} \text{B}_{3\times3}$ is a box filter. $\mathbbm{1}$ is an indicator function. Finally, the robust mask is defined as:  
\begin{equation}
    \mathcal{M}_{final} = \mathcal{M}_{global} \cup \mathcal{M}_{local}.
\end{equation}By utilizing the robust mask $\mathcal{M}_{final}$,  we more accurately capture ambiguous distractors independent of color and semantic similarity compared to prior methods. During optimization, the total loss is defined as:
\begin{equation}
\begin{aligned}
\mathcal{L}_{\mathrm{GS}}
&=(1-\lambda)\,\mathcal{M}_{\mathrm{final}} \odot \mathcal{L}_1(I_{pred}, I_{gt}) \\
&\quad + \lambda\,\mathcal{M}_{\mathrm{final}} \odot \mathcal{L}_{\mathrm{D\text{-}SSIM}}(I_{pred}, I_{gt}).
\end{aligned}
\end{equation}

\subsection{Entropy-aware density control}
\input{tab_fig/fig_method2}
\label{subsec:Entropy-awareDensityControl}
\noindent \textbf{Entropy-aware gradient for density control.} Most prior methods \cite{park2026forestsplats, fu2025robustsplat} utilize the derivation of $\mathcal{L}_{GS}$ as a positional gradient in density control as:
\begin{equation}
\frac{\partial \mathcal{L}_{\text{GS}}}{\partial\mu} =\frac{\partial \mathcal{L}_{\text{GS}}}{\partial C} \frac{\partial C}{\partial\mu}=\frac{\partial \mathcal{L}_{\text{GS}}}{\partial C}\left(\frac{\partial C}{\partial c} \frac{\partial c}{\partial \mu}+\frac{\partial C}{\partial \alpha} \frac{\partial \alpha}{\partial \mu}\right).
\label{fig:eq_problem}
\end{equation}
Since the derivative of $\mathcal{L}_{GS}$ primarily focuses on photometric error and density, aligning Gaussians during the density control remains challenging. This is particularly evident in regions where distractors share similar color or semantic features with the static background. Motivated by the limitations of photometric gradients, we propose the simple yet effective entropy-aware density control. This approach adaptively clones or splits Gaussians by incorporating entropy into the positional gradients, as shown in Fig. \ref{fig:entropy-aware-densify} and Eq. \ref{eq:loss_id}. To facilitate the cloning and splitting process in ambiguous regions, we derive the generation of new Gaussians based on the gradient of $\mathcal{L}_{id}$ as follows:
\begin{equation}
\label{eq:loss_id}
\frac{\partial \mathcal{L}_{id}}{\partial \mu} = \frac{\partial \mathcal{H}}{\partial f_{pred}} \frac{\partial f_{pred}}{\partial D_\theta} \frac{\partial D_\theta}{\partial \hat I} \frac{\partial \hat I}{\partial \mu}, 
\end{equation} where $\mathcal{L}_{id}$ is defined as $\mathcal{L}_{id} = CE(\psi_{id}(D_{\theta}(\hat{I})) ,S(I^{gt}))$. Specifically, in contrast to ForestSplats \cite{park2026forestsplats}, which relies on uncertainty embeddings and photometric error, our method effectively aligns Gaussians within the same identity by leveraging entropy. This approach mitigates artifact issues without requiring additional embeddings.
\input{tab_fig/fig_tab_ambi}

\noindent \textbf{Merging process for overlapping Gaussians.}
After densification, we cluster and select pairs to merge Gaussians considering the overlap ratio and color similarity. This process effectively reduces redundant Gaussians and mitigates artifacts from densification. The overlap ratio $\mathcal{T}_{over}$ between the i-th and j-th neighboring Gaussians is defined as: 
\begin{equation}
    \mathcal{T}_{over} = \frac{\exp \left( -\frac{1}{2}(\mu_{i} - \mu_{j})^{\top}(\Sigma_{sum})^{-1}(\mu_{i} - \mu_{j}) \right)}{\sqrt{(2\pi)^{3}|\Sigma_{sum}|}},
\end{equation}
where $\Sigma_{sum}$ is computed as $\Sigma_i+\Sigma_j$. For each clustered Gaussian, a new Gaussian with attributes $(\mu_{k}^{*}, c_{k}^{*}, \alpha_{k}^{*}, \Sigma_{k}^{*})$ is defined by aggregating the Gaussians within each cluster in a weighted-based manner as:
\begin{equation}
\begin{gathered}
\mu_k^* = \frac{\sum_{i=1}^{N} \alpha_i \mu_i}{\sum_{i=1}^{N} \alpha_i}, \quad
c_k^* = \frac{\sum_{i=1}^{N} \alpha_i c_i}{\sum_{i=1}^{N} \alpha_i}, \quad
\alpha_k^* = \frac{\sum_{i=1}^{N} \alpha_i}{N}, \\
\Sigma_k^* = 
\frac{\sum_{i=1}^{N} \alpha_i \left(\Sigma_i + (\mu_i - \mu_k^*)(\mu_k^* - \mu_i)^\top \right)}
{\sum_{i=1}^{N} \alpha_i},
\end{gathered}
\end{equation}
where $N$ denotes the Gaussian number of clustered set. This merging process align and reduces redundant Gaussians. 
\subsection{Regularization and Optimization}
\label{subsec:Regularization}
\noindent \textbf{Consistency regularization.} Since the extracted instance masks \cite{meng2025zero} includes minor errors, we introduce the consistency regularization to rectify incorrect instance masks and reduce the variance of entropy in segments, as follows: 
\begin{equation}
    \mathcal{L}_{cr} = \frac{1}{K} \sum_{j=1}^{K} \frac{1}{N^2} \sum_{p \in \mathcal{M}^j}^{N} \sum_{q \in \mathcal{M}^j}^{N} D_{KL}( \sigma(\mathcal{E}_i(p)) \,\|\, \sigma(\mathcal{E}_i(q))).
\end{equation} This encourages all pixels within each segment to represent a consistent identity, promoting precise transient masks.

\input{tab_fig/fig_tab_diverse}

\vspace{0.2\baselineskip}
\noindent \textbf{Optimization.} Finally, the total loss is defined as:
\begin{equation}
\mathcal{L}_{total} =  \mathcal{L}_{GS} +\mathcal{L}_{id} + \alpha_1\mathcal{L}_{cr},
\end{equation}where $\alpha_1$ is 0.2. During optimization, we periodically leverage the entropy-aware density control to align Gaussians. Furthermore, we also utilize the derivation of $\mathcal{L}_{cr}$ to reflect entropy, cloning or splitting Gaussians, and effectively aligning Gaussians in real-world scenarios.

%% file: tab_fig/fig_method.tex
\begin{figure}
    \centering
    \includegraphics[width=\linewidth]{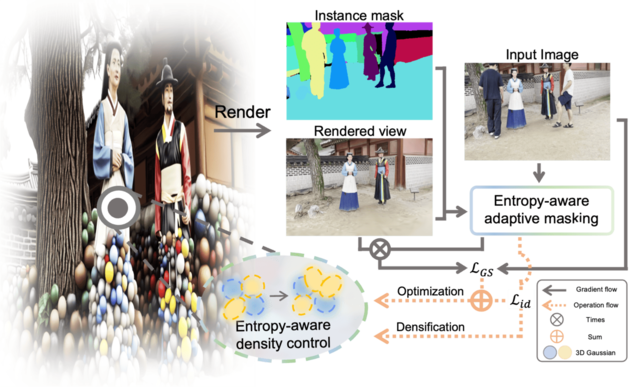}
    \vspace{-2.3em}
    \caption{Overview of RefineSplat. We construct Entropy-aware adaptive masks by leveraging statistics of entropy and instance masks. Furthermore, we effectively align new Gaussians by utilizing entropy magnitude as a positional gradient and merging Gaussians across real-world scenarios.}
    \label{fig:methods}
    \vspace{-1.5em}
\end{figure}

%% file: tab_fig/fig_method2.tex
\begin{figure}[!t]
    \centering
    \includegraphics[width=\linewidth]{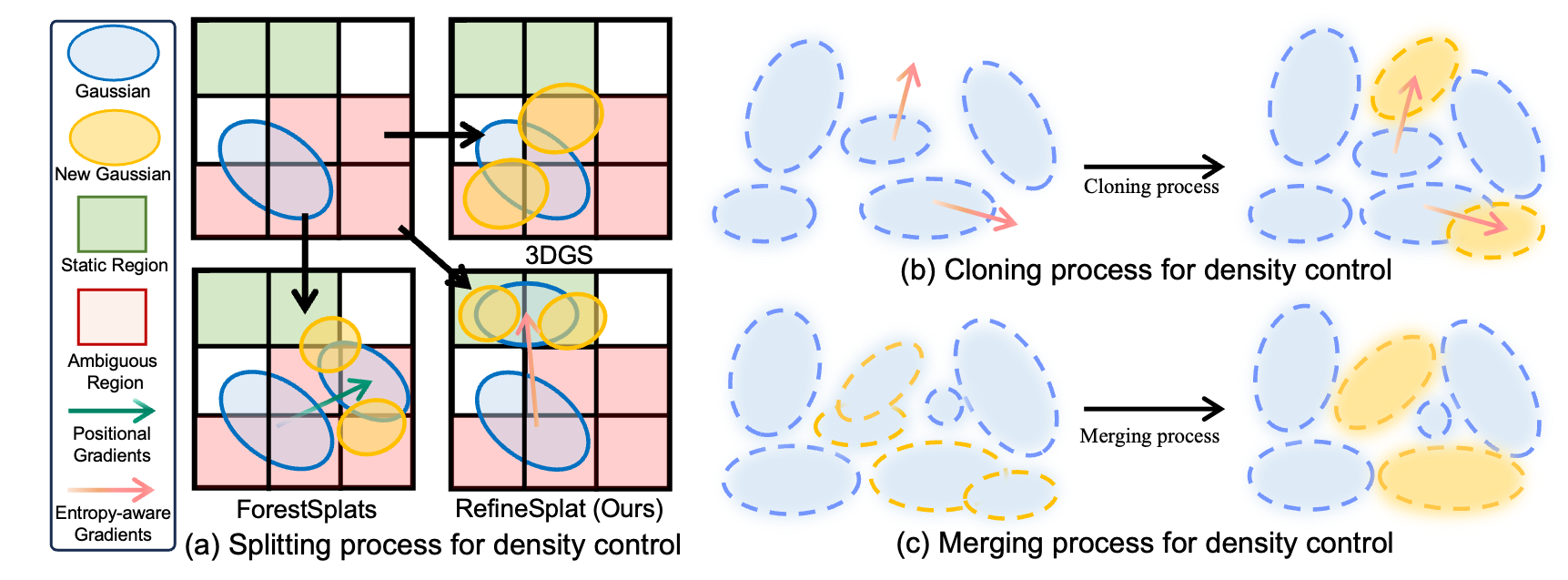}
    \vspace{-2em}
    \caption{Entropy-aware density control. (a) Compared to existing methods, we split Gaussians utilizing entropy-aware gradients, aligning new Gaussians in static regions. (b) For the clone process, we generate new Gaussians following entropy-aware gradients. (c) We merge overlapping Gaussians considering color similarity, reducing redundant Gaussians and artifact issues.} 
    \label{fig:entropy-aware-densify}
    \vspace{-1.5em}
\end{figure}

%% file: tab_fig/fig_tab_ambi.tex
\begin{table*}[!t]
	\centering
    \renewcommand{\arraystretch}{1.05}
	\resizebox{\linewidth}{!}{
		\begin{tabular}{ccccccccccccccccccc}
			\toprule
			\multirow{2}{*}{Method} & \multicolumn{3}{c}{Humanoid}& \multicolumn{3}{c}{Porter} &  \multicolumn{3}{c}{Lounge} & \multicolumn{3}{c}{Bust} & \multicolumn{3}{c}{Jockey} & \multicolumn{3}{c}{Statuette} \\
			& {\scriptsize PSNR$\uparrow$} & {\scriptsize SSIM$\uparrow$} & {\scriptsize LPIPS$\downarrow$} & {\scriptsize PSNR$\uparrow$} & {\scriptsize SSIM$\uparrow$} & {\scriptsize LPIPS$\downarrow$} & {\scriptsize PSNR$\uparrow$} & {\scriptsize SSIM$\uparrow$} & {\scriptsize LPIPS$\downarrow$}  & {\scriptsize PSNR$\uparrow$} & {\scriptsize SSIM$\uparrow$} & {\scriptsize LPIPS$\downarrow$}  & {\scriptsize PSNR$\uparrow$} & {\scriptsize SSIM$\uparrow$} & {\scriptsize LPIPS$\downarrow$}& {\scriptsize PSNR$\uparrow$} & {\scriptsize SSIM$\uparrow$} & {\scriptsize LPIPS$\downarrow$}  \\
                        \midrule
            \midrule
			Mip-Splatting \cite{yu2024mip} 
            & 15.15 & 0.732 & 0.279
            & 19.12 & 0.661 & 0.194
            & 19.12 & 0.661 & 0.194
            & 20.66 & \metrictablethird{0.688} & 0.251
            & 15.12 & 0.407 & 0.499
            & 18.10 & 0.664 & 0.333 \\
			GS-W \cite{zhang2024gaussian}     
            & 18.31 & 0.759 & \metrictablesecond{0.251}  
            & 19.65 & 0.685 & 0.260 
            & 21.21 & 0.885 & 0.206
            & 20.40 & 0.552 & 0.461 
            & 13.33 & 0.451 & 0.481 
            & 19.08 & 0.625 & 0.308 \\
			NexusSplats \cite{tang2024nexussplats} & 16.56 & 0.776 & 0.351 & 20.51 & 0.701 & 0.291 & 22.55 & 0.872 & \metrictablethird{0.189} & 19.81 & 0.509 & 0.474 & 15.41 & 0.416 & 0.464 & 18.89 & 0.603 & 0.348 \\
			WildGaussians \cite{kulhanek2024wildgaussians} & 19.02 & 0.801 & 0.355 & 18.54 & 0.664 & 0.272 & 18.52 & 0.825 & 0.266 & 19.59 & 0.457 & 0.585 & 13.46 & 0.395 & 0.578 & 18.03 & 0.549 & 0.441 \\
			DroneSplat \cite{tang2025dronesplat} 
            & 14.89 & 0.695 & 0.315 
            & \metrictablesecond{22.45} & \metrictablesecond{0.819} & \metrictablesecond{0.129} 
            & 22.68 & 0.855 & 0.199
            & \metrictablesecond{22.40} & \metrictablesecond{0.694} & \metrictablesecond{0.243} 
            & \metrictablethird{16.84} &  \metrictablesecond{0.553} & \metrictablefirst{0.284} 
            & \metrictablesecond{20.16} & \metrictablesecond{0.729} & \metrictablesecond{0.197} \\
			RobustSplat \cite{fu2025robustsplat} 
            & \metrictablethird{19.10} & \metrictablethird{0.812} & 0.309  
            & 21.20 & 0.706 & 0.327 
            & \metrictablesecond{22.82} & \metrictablesecond{0.895} & \metrictablesecond{0.172} 
            & \metrictablethird{21.20} & 0.529 & 0.472 
            & \metrictablesecond{17.76} & \metrictablethird{0.478} & 0.496  
            & 20.01 & 0.614 & 0.383 \\
            AsymGS \cite{li2026robust}
            & \metrictablesecond{19.53} & \metrictablesecond{0.828} & \metrictablethird{0.272} & \metrictablethird{22.38} & \metrictablethird{0.810} & \metrictablethird{0.131} & \metrictablethird{21.53} & \metrictablethird{0.889} & 0.198 & 21.26 & 0.696 & \metrictablethird{0.245} & 17.02 & 0.489 & \metrictablethird{0.331} & \metrictablethird{20.12} & \metrictablethird{0.714} & \metrictablethird{0.237} \\
			RefineSplat(Ours) 
            & \metrictablefirst{19.71} & \metrictablefirst{0.841} & \metrictablefirst{0.173}  
            & \metrictablefirst{24.30} & \metrictablefirst{0.873} & \metrictablefirst{0.128} 
            & \metrictablefirst{23.14} & \metrictablefirst{0.902} & \metrictablefirst{0.143} 
            & \metrictablefirst{22.91} & \metrictablefirst{0.708} & \metrictablefirst{0.225} 
            & \metrictablefirst{18.45} & \metrictablefirst{0.588} & \metrictablesecond{0.308} 
            & \metrictablefirst{22.04} & \metrictablefirst{0.758} & \metrictablefirst{0.194} \\
			\hline
		\end{tabular}
	}
    \vspace{-1em}
    \caption{Quantitative results on the Ambiguous wild dataset. Performance is highlighted by color from \metrictablethird{third} \metrictablesecond{to} \metrictablefirst{first}.}
	\label{tab:ambi_quanti}
    \vspace{-1.3em}
\end{table*}
\begin{figure*}[!t]
    \centering
    \includegraphics[width=\textwidth]{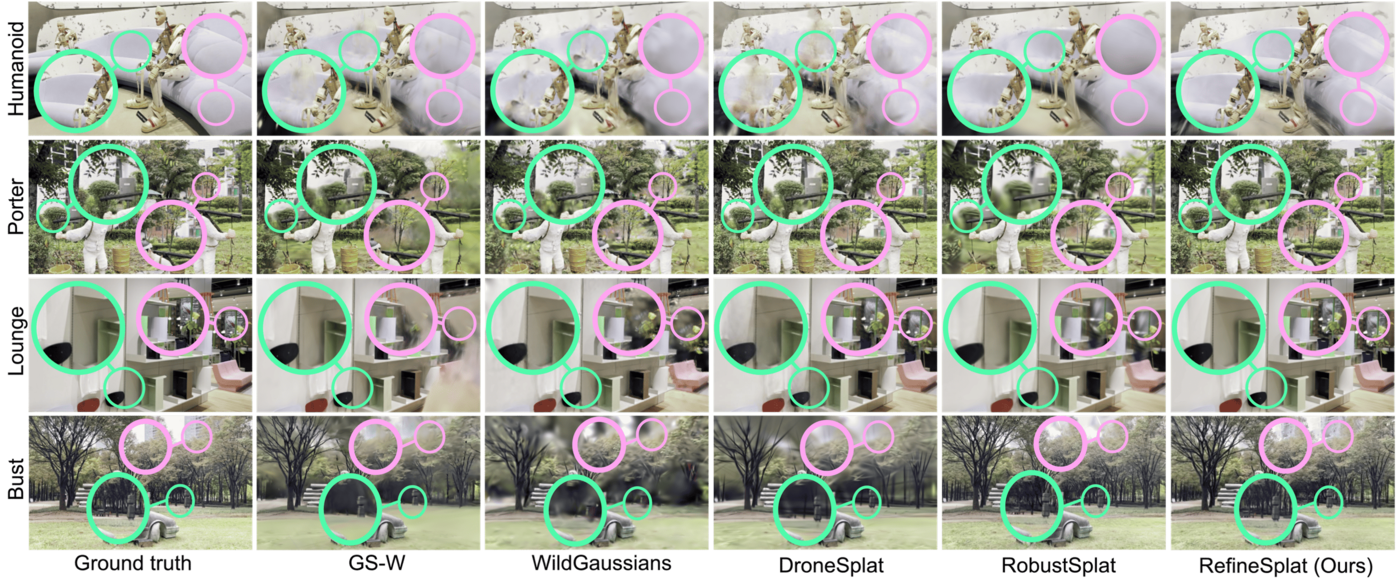}
    \vspace{-2.2em}
    \caption{Qualitative results from novel-view synthesis on the Ambiguous wild dataset.} 
    \label{fig:ambi}
    \vspace{-1.5em}
\end{figure*}

%% file: tab_fig/fig_tab_diverse.tex
\begin{table*}[!t]
	\centering
    \renewcommand{\arraystretch}{0.9}
	\resizebox{\linewidth}{!}{
		\begin{tabular}{ccccccccccccc}
			\toprule
			\multirow{2}{*}{Method} & \multicolumn{3}{c}{NeRF On-the-go}& \multicolumn{3}{c}{Photo Tourism} &  \multicolumn{3}{c}{Drone Imagery} & \multicolumn{3}{c}{Average} \\
			& {\scriptsize PSNR$\uparrow$} & {\scriptsize SSIM$\uparrow$} & {\scriptsize LPIPS$\downarrow$} & {\scriptsize PSNR$\uparrow$} & {\scriptsize SSIM$\uparrow$} & {\scriptsize LPIPS$\downarrow$} & {\scriptsize PSNR$\uparrow$} & {\scriptsize SSIM$\uparrow$} & {\scriptsize LPIPS$\downarrow$}  & {\scriptsize PSNR$\uparrow$} & {\scriptsize SSIM$\uparrow$} & {\scriptsize LPIPS$\downarrow$}    \\
            \midrule
			3D-GS \cite{kerbl20233d} & 19.30 & 0.668 & 0.252 & 18.03 & 0.814 & 0.183 & 17.12  & \metrictablethird{0.622} & 0.301 & 18.15 & 0.701 & 0.245 \\
			HA-NeRF \cite{chen2022hallucinated} & 17.65 & 0.504 & 0.488 & 21.41 & 0.789 & 0.177 & 20.30 & 0.589 & 0.286 & 19.79 & 0.627 & 0.317 \\
			Mip-Splatting \cite{yu2024mip} & 19.60 & 0.676 & 0.232 & 18.70 & 0.838 & \metrictablesecond{0.170} & 18.50 & 0.573 & 0.272 & 18.93 & 0.696 & 0.225 \\
			Splatfacto-W \cite{xu2024splatfacto} & 18.96 & 0.621 & 0.332 & \metrictablethird{23.88} & \metrictablesecond{0.857} & \metrictablethird{0.172} & 19.81 & 0.576 & \metrictablesecond{0.232} & 20.88 & 0.685 & 0.261 \\ 
			GS-W \cite{zhang2024gaussian} & 19.76 & 0.680 & 0.276 & 20.98 & 0.815 & 0.217 & 19.87 & 0.560 & 0.257 & 20.20 & 0.685 & 0.250   \\
			WildGaussians \cite{kulhanek2024wildgaussians} & 22.14 & 0.746 & 0.166 & \metrictablesecond{24.65} & \metrictablethird{0.850} & 0.179 & 19.89 & 0.539 & 0.311 & \metrictablesecond{22.15} & \metrictablethird{0.712} & \metrictablethird{0.218} \\
			DroneSplat \cite{tang2025dronesplat} & \metrictablethird{22.71} & \metrictablethird{0.771} & \metrictablefirst{0.136} & 17.81 & 0.780 & 0.190 & \metrictablethird{20.82} & \metrictablesecond{0.624} & \metrictablefirst{0.188} & 20.44 & \metrictablesecond{0.726} & \metrictablefirst{0.171} \\
			RobustSplat \cite{fu2025robustsplat} & \metrictablesecond{23.21} & \metrictablesecond{0.817} & \metrictablesecond{0.148} & 18.41 & 0.649 & 0.350 & \metrictablesecond{21.17} & 0.575 & 0.468 & \metrictablethird{20.93} & 0.680 & 0.322 \\
			RefineSplat(Ours) & \metrictablefirst{23.44} & \metrictablefirst{0.820} & \metrictablethird{0.154} & \metrictablefirst{25.22} & \metrictablefirst{0.870} & \metrictablefirst{0.152} & \metrictablefirst{21.86} & \metrictablefirst{0.658} & \metrictablethird{0.234} & \metrictablefirst{23.51} & \metrictablefirst{0.782} & \metrictablesecond{0.189} \\ 
			\hline
		\end{tabular}
	}
    \vspace{-1em}
    \caption{Quantitative average results on diverse real-world datasets. Performance is highlighted by color, from \metrictablethird{third} \metrictablesecond{to} \metrictablefirst{first}.}
    \label{tab:sota_diverse}
    \vspace{-1.3em}
\end{table*}

%% file: section/experiments.tex
\input{tab_fig/fig_abl_mask_compare}
\input{tab_fig/fig_abl_mask}
\section{Experiments}
\noindent \textbf{Implementation details.} RefineSplat is developed based on DroneSplat \cite{tang2025dronesplat} and RobustSplat \cite{fu2025robustsplat}. All methods are trained following their own default settings. We train our method with 30K iterations, and the same learning rate as 3DGS \cite{kerbl20233d} using RTX 4090 GPU on the Ambiguous wild and Nerf On-the-go \cite{ren2024nerf} datasets. For the PhotoTourism \cite{snavely2006photo} dataset, we train our method with 200k iterations. We also use the Adam optimizer \cite{kingma2014adam} with weight decay. In addition, we utilize the frozen segmentation model \cite{cheng2023tracking} to utilize instance masks. Moreover, the semantic features used by the MLPs are extracted from frozen DINOv2 \cite{oquab2023dinov2}, using pre-trained weights from the distilled ViT-S/14 model. We also apply a down-sampling factor of 8 on the NeRF On-the-go and Ambiguous wild, following prior methods \cite{fu2025robustsplat, kulhanek2024wildgaussians}. We train 2-layer MLPs $\psi_{id}$ using extracted instance masks, which takes just a few minutes before training the Gaussian field. Note that although our method can optimize the MLPs and the Gaussian field simultaneously, we first train the MLPs to simplify the overall pipeline.

\input{tab_fig/fig_abl_mask_densi}
\input{tab_fig/fig_abl_depth_opacity}

\vspace{0.2\baselineskip}
\noindent \textbf{Datasets, metrics, and baselines.} We evaluate RefineSplat on three datasets: Photo Tourism \cite{snavely2006photo}, NeRF On-the-go \cite{ren2024nerf}, and Drone Imagery \cite{tang2025dronesplat}. We propose the Ambiguous wild dataset that contains 18 casually captured images, including 11 outdoor and 7 indoor scenes. To evaluate RefineSplat, we use PSNR, SSIM \cite{wang2004image}, and LPIPS \cite{zhang2018unreasonable} as metrics. We compare RefineSplat with prior work. (1) Residual-based masking: Ha-NeRF \cite{chen2022hallucinated}, 3DGS \cite{kerbl20233d}, Mip-Splatting \cite{yu2024mip}, Splatfacto-W \cite{xu2024splatfacto}, GS-W \cite{zhang2024gaussian}; (2) Semantic-based masking: WildGaussians \cite{kulhanek2024wildgaussians}, NexusSplats \cite{tang2024nexussplats}, RobustSplat \cite{fu2025robustsplat}; (3) Heuristic-based masking: DroneSplat \cite{tang2025dronesplat}. More baselines are included in the supplementary material.

\noindent \textbf{Experimental analysis.} For qualitative analysis, we show the rendering results of several methods \cite{zhang2024gaussian, kulhanek2024wildgaussians, tang2025dronesplat, fu2025robustsplat} as shown in Fig. \ref{fig:ambi} 
Furthermore, we also provide quantitative results as shown in Tab. \ref{tab:ambi_quanti} and Tab. \ref{tab:sota_diverse}. Compared to our method, RobustSplat \cite{fu2025robustsplat} and WildGaussians \cite{kulhanek2024wildgaussians}, which only utilize extracted semantic features  to capture distractors, struggle to tackle distractors that have similar semantic levels to static elements. GS-W \cite{zhang2024gaussian} also only focuses on the residual map, which makes it hard to discern distractors that possess a similar color to the static background. DroneSplat \cite{tang2025dronesplat} leverages heuristic invariant masks obtained from SAM \cite{kirillov2023segment} before training without error correction, which heavily depend on strict boundaries. In contrast to prior work, our method leverages entropy and instance masks to robustly capture ambiguous distractors.

%% file: tab_fig/fig_abl_mask_compare.tex
\begin{figure}[!t]
    \centering
    \includegraphics[width=\linewidth]{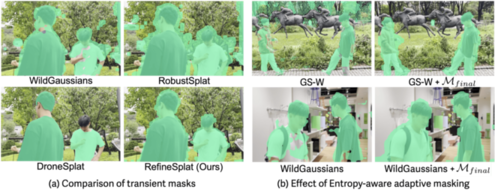}
    \vspace{-2em}
    \caption{Analysis of the entropy-aware adaptive masking.}
    \label{fig:transient_mask}
    \vspace{-1.1em}
\end{figure}
\begin{table}[!t]
    \centering
     \resizebox{\linewidth}{!}{
    \begin{tabular}{cccccccc}
    \toprule
    \multirow{2}{*}{Method} & \multirow{2}{*}{w/ $\mathcal{M}_{final}$ } & \multicolumn{3}{c}{Threshold} & \multicolumn{3}{c}{Humanoid} \\
     & & $\mathcal{T}_{\text{hard}}$ & $\mathcal{T}_{\text{statistics}}$  & $\mathcal{T}_{\text{ent}}$  & PSNR $\uparrow$ & SSIM $\uparrow$ & LPIPS $\downarrow$  \\
     \midrule
     \midrule
     \multirow{4}{*}{\raisebox{-1.25ex}{GS-W \cite{zhang2024gaussian}}}  & - & \boldcheckmark & - & - &  $20.31$ & $0.868$ &  $0.251$ \\
     & \boldcheckmark & \boldcheckmark & - & - &  $21.17$ & $0.874$ & $0.243$ \\
     & \boldcheckmark & - & \boldcheckmark & - & $21.82$ &$0.877$ & $0.229$ \\ 
     & \boldcheckmark & - & - & \boldcheckmark & $22.15$ & $0.881$ &  $0.206$ \\
     \midrule
     \multirow{4}{*}{\raisebox{-1.25ex}{\;\;WildGaussians \cite{kulhanek2024wildgaussians}}}  & - & \boldcheckmark & - & - &  $19.10$ & $0.801$ & $0.355$ \\
     & \boldcheckmark & \boldcheckmark & - & - & $19.84$ & $0.847$ & $0.302$ \\
     & \boldcheckmark & - & \boldcheckmark & - &  $20.17$ & $0.859$ & $0.289$ \\
     & \boldcheckmark & - & - & \boldcheckmark & $20.31$ & $0.861$ & $0.283$ \\
     \bottomrule
    \end{tabular}}
    \vspace{-1em}
    \caption{Ablation of the threshold with existing methods.}
    \label{tab:entropy-aware_plug}
    \vspace{-1.3em}
\end{table}

%% file: tab_fig/fig_abl_mask.tex
\begin{figure}[!t]
    \centering
    \includegraphics[width=\linewidth]{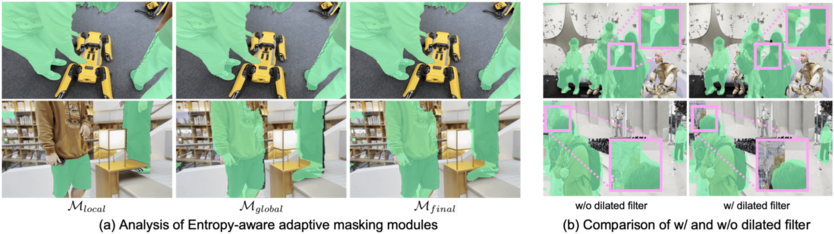}
    \vspace{-2em}
    \caption{Analysis of the entropy-aware adaptive masking modules. (a) Analysis of Entropy-aware adaptive mask modules. (b) Comparison of w/ and w/o dilated filter.} 
    \label{fig:mask_module_dilated}
    \vspace{-1.5em}
\end{figure}

%% file: tab_fig/fig_abl_mask_densi.tex
\begin{table}[!t]
    \centering
    \footnotesize
    \renewcommand{\arraystretch}{0.7}
    \resizebox{\linewidth}{!}{
    \begin{tabular}{cccccc}
        \toprule
         & \multicolumn{2}{@{\hspace{-0.64em}}c}{Entropy-aware adaptive mask} 
         & \multicolumn{3}{c}{Humanoid}  \\ 
         &  \;\; $\mathcal{M}_{local}$ & $\mathcal{M}_{global}$ & PSNR $\uparrow$  
         & SSIM $\uparrow$ & LPIPS $\downarrow$ \\  
        \midrule
        \midrule
        (a) &\multicolumn{2}{c}{Baselines \cite{fu2025robustsplat}} & $20.19$ & $0.854$ & $0.275$\\
        (b) & \boldcheckmark & - & $21.33$ & $0.873$ & $0.198$ \\
        (c) & - & \boldcheckmark & $20.46$ & $0.851$ & $0.279$ \\
        (d) & \boldcheckmark & \boldcheckmark & $22.11$ & $0.897$ & $0.173$  \\
        \bottomrule
    \end{tabular}}
    \hspace*{1.8mm}
    \includegraphics[width=\linewidth]{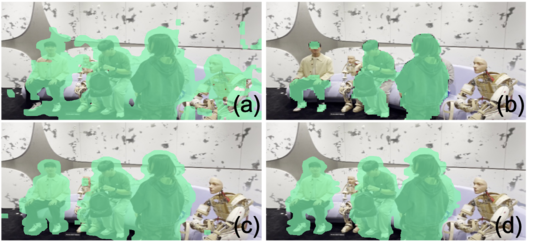}
    \vspace{-2.7em}
    \caption{Ablation of the entropy-aware adaptive masking.}
    \label{tab:mask_complement}
    \vspace{-1.6em}
\end{table}
\begin{table}[!t]
    \centering
    \resizebox{\linewidth}{!}{
    \begin{tabular}{ccccccc}
        \toprule
         & \multicolumn{2}{@{\hspace{-0.64em}}c}{Entropy-aware density control} 
         & \multicolumn{3}{c}{Jewel} & 
          \multirow{2}{*}{Memory (MB)} \\ 
         &  Entropy & Merging & PSNR $\uparrow$ 
         & SSIM $\uparrow$ & LPIPS $\downarrow$ & \\  
        \midrule
        \midrule
        (a) &\multicolumn{2}{c}{Baselines \cite{fu2025robustsplat}} & $26.62$ & $0.894$ & $0.103$ & $231.51$ \\
        (b) & \boldcheckmark & - & $27.43$ & $0.911$ & $0.095$ &  $186.09$ \\
        (c) & - & \boldcheckmark & $26.79$ & $0.898$ & $0.097$ & $142.86$ \\
        (d) & \boldcheckmark & \boldcheckmark & $27.88$ & $0.918$ & $0.089$  & $153.37$ \\
        \bottomrule
    \end{tabular}}
    \includegraphics[width=\linewidth]{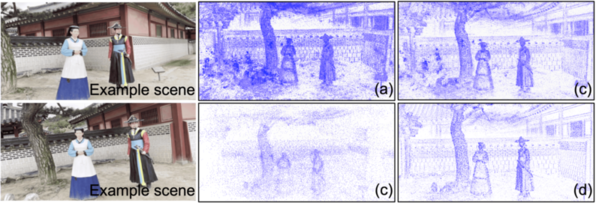}  
    \vspace{-2.2em}
    \caption{Analysis of the entropy-aware density control.}
    \label{tab:density_control}
    \vspace{-1.5em}
\end{table}

%% file: tab_fig/fig_abl_depth_opacity.tex
\begin{figure}[!t]
    \centering
    \includegraphics[width=\linewidth]{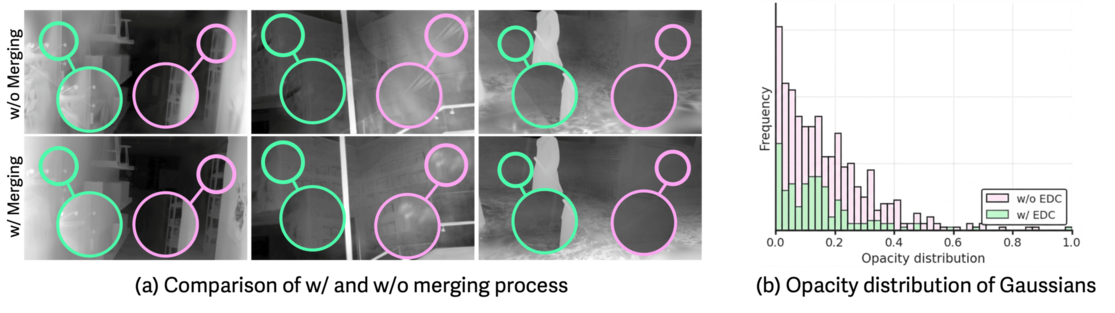}
    \vspace{-2em}
    \caption{Analysis of the Entropy-aware density control. (a) Comparison of depth consistency w/ and w/o merging process. (b) Comparison of opacity distribution of Gaussians w/ and w/o entropy-aware density control (EDC).} 
    \label{fig:entropy_densify}
    \vspace{-1.3em}
\end{figure}
\begin{table}[!t]
    \centering
    \resizebox{1.0\linewidth}{!}{
    \begin{tabular}{cccccc}
    \toprule
         Scene  & Method &  Memory  & Scene &  Method & Memory\\ 
        \midrule 
        \midrule
        \multirow{3}{*}{{\shortstack{Humanoid}}} & RobustSplat \cite{fu2025robustsplat} &  110.01 & \multirow{3}{*}{Patio} & RobustSplat \cite{fu2025robustsplat} &86.68 \\
        & DroneSplat  \cite{tang2025dronesplat} &  51.32 & & DroneSplat \cite{tang2025dronesplat}   & 114.13  \\
        & RefineSplat (Ours) & 43.24 & & RefineSplat (Ours) & 86.61 \\
        \midrule 
        \midrule
        \multirow{3}{*}{\shortstack{Spot}} & RobustSplat \cite{fu2025robustsplat} & 64.39 & \multirow{3}{*}{\shortstack{Sacre\\Coeur}} & RobustSplat \cite{fu2025robustsplat} & 82.15 \\
        & DroneSplat \cite{tang2025dronesplat} & 79.24 & & DroneSplat \cite{tang2025dronesplat} &  77.12 \\
        &  RefineSplat (Ours) & 25.80 & & RefineSplat (Ours) & 31.26 \\
    \bottomrule
    \end{tabular}}
    \vspace{-1em}
    \caption{Analysis of memory usage in real-world scenarios.}
    \label{tab:memory_size}
    \vspace{-1.3em}    
\end{table}
\begin{table}[!t]
    \centering
    \resizebox{\linewidth}{!}{
    \begin{tabular}{ccccccccc}
    \toprule
    & PUP  \;& UAD \; & EDC (Ours) &  PSNR $\uparrow$ &SSIM $\uparrow$ &LPIPS $\downarrow$ & MB \\
    \midrule
    \midrule
    (a) & \checkmark&  & & 20.77  & 0.704 & 0.283   & 42.1   \\
    (b) & & \checkmark & & $21.16$  & $0.719$ & $0.271$  & $39.5$ \\
    (c) & & & \checkmark & $23.24$  & $0.743$  & $0.228$  & $30.5$ \\
    \bottomrule
    \end{tabular}}
    \includegraphics[width=\linewidth]{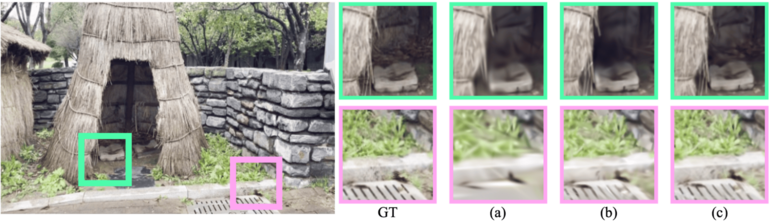} 
    \vspace{-2.2em}
    \caption{Comparison of the density control strategy.}
    \label{tab:density_sota}
    \vspace{-1.3em}
\end{table}

%% file: section/ablation.tex
\input{tab_fig/fig_abl_control_sota}

\section{Ablation Study}
\noindent \textbf{Effectiveness of the entropy-aware adaptive masking.} Although prior work primarily focus on photometric error or semantic similarity to identify transient elements, as presented in Fig. \hyperref[fig:transient_mask]{\ref{fig:transient_mask}-(a)}, the entropy-aware adaptive masking  distinguishes ambiguous distractors from static scenes by leveraging both semantic levels and instance levels. Furthermore, the entropy-aware adaptive masking can function as a plug-and-play module, as shown in Fig. \hyperref[fig:transient_mask]{\ref{fig:transient_mask}-(b)} and Tab. \ref{tab:entropy-aware_plug}. Although utilizing thresholds based on statistical measures (\eg., mean, variance, percentile) $\mathcal{T}_{statistis}$ or fixed values $\mathcal{T}_{hard}$ show competitive results, it is struggle to capture diverse distractors in skew distributions, as depicted in Fig. \ref{fig:method_motivation}. However, adding the entropy-aware adaptive masking to GS-W \cite{zhang2024gaussian} and WildGaussians \cite{kulhanek2024wildgaussians} effectively captures transient elements, showing robustness for $\mathcal{L}_{ent}$ in ambiguous scenarios. We also illustrate how the entropy-based threshold works, as shown in Fig. \ref{fig:method_motivation}. We also observe that leveraging global threshold $\mathcal{T}_{glb}$ effectively constructs masks in capturing distractors compared to a fixed threshold $\mathcal{T}_{hard}$, as depicted in Tab. \ref{tab:treshold_glb_mask}. Moreover, we observe that synthesizing the entropy-aware adaptive global and local masks plays a key role in creating robust transient masks by complementing their respective limitations, as shown in Tab. \ref{tab:mask_complement} and Fig. \hyperref[fig:mask_module_dilated]{\ref{fig:mask_module_dilated}-(a)}. We observe that leveraging a box filter, following prior work \cite{sabour2024spotlesssplats, sabour2023robustnerf}, considers neighboring context to capture distractors, as depicted in Fig. \hyperref[fig:mask_module_dilated]{\ref{fig:mask_module_dilated}-(b)}. Through extensive analysis, we demonstrate that the entropy-aware adaptive masking precisely identifies ambiguous distractors.

\vspace{0.2\baselineskip}
\noindent \textbf{Efficiency of the entropy-aware density control.} As shown in Tab. \ref{tab:density_control}, our strategy confirms memory efficiency compared to the baselines and arranges Gaussian anchors without artifacts. Furthermore, we observe that using the entropy-aware density control (EDC) reduces redundant Gaussians, leaving the number of high opacity Gaussians, as shown in Fig. \hyperref[fig:entropy_densify]{\ref{fig:entropy_densify}-(a)} and Fig. \hyperref[fig:entropy_densify]{\ref{fig:entropy_densify}-(b)}. Moreover, to rigorously evaluate the effectiveness of the entropy-aware density control, we compare our methods with prior methods \cite{hanson2025pup, park2026forestsplats, fu2025robustsplat}, as shown in Tab. \ref{tab:density_sota}, and Fig. \ref{fig:depth_sota}. Although prior methods also show impressive results by utilizing photometric-based gradients, these methods struggle to align Gaussians in ambiguous scenarios. However, our method considers both entropy to align Gaussians, mitigating artifact issues. In addition, our method shows memory efficiency, maintaining visual quality, as shown in Tab. \ref{tab:memory_size}.

\input{tab_fig/fig_sensitivity}
\input{tab_fig/fig_abl_regularization_alpha}

\vspace{0.2\baselineskip}
\noindent \textbf{Analysis of the each module.} As shown in Fig.~\ref{fig:mask_correction}, we confirm that the consistency regularization rectifies the error in instance masks. In addition, we examine the sensitivity to different hyperparameter settings, as shown in Tab. \ref{main_tab:sensitive_cr}. The results indicate that the consistency regularization is robust to hyperparameter variations. Moreover, to analyze each module, we perform an ablation study to demonstrate the effectiveness: Entropy-aware adaptive masking (EAM), Entropy-aware density control (EDC), and Consistency regularization (CR), as shown in Tab. \ref{tab:impact_of_module}. The analysis shows that entropy-aware adaptive masking captures ambiguous elements. We also analyze how sensitive it is to patch sizes and DINO variants on the Patio scene, as shown in Tab. \ref{tab:dino_patch_size} and Tab. \ref{tab:dino_variant}. Moreover, we continue to test dilation filter sizes, as shown in Tab. \ref{tab:dilated_filter}. We show more experiments for color threshold values for merging, as shown in Tab. \ref{rebuttal_tab:treshold}. By observing them, we set a dilation filter size of 5. 

\vspace{0.2\baselineskip}
\noindent \textbf{Discussion on the Ambiguous wild dataset.} We capture 18 natural scenes exemplifying different types of distractors, including varying ratios of distractors (from 5\% to over 60\%) and ambiguity, as shown in Fig. \hyperref[fig:data_description]{\ref{fig:data_description}-(b)}. Furthermore, as illustrated in Fig. \ref{fig:motivation}, to evaluate the ambiguity of the Ambiguous wild dataset, we consider three criteria: residual error, cosine similarity, and human evaluation (User Study). Moreover, we observe that leveraging COLMAP \cite{schonberger2016structure} on the Ambiguous wild dataset, where distractors and static elements are visually or semantically similar, tends to produce incorrect matching points, which hinders the initialization of Gaussians, as shown in Fig. \hyperref[fig:data_description]{\ref{fig:data_description}-(a)}.
\input{tab_fig/tab_feature_sensitivity}
\input{tab_fig/tab_sensitivitiy_parameters}

%% file: tab_fig/fig_abl_control_sota.tex
\begin{table}[]
    \centering
    \setlength{\tabcolsep}{9.5pt}
    \renewcommand{\arraystretch}{0.9}
    \begin{tabular}{ccccc}
        \toprule
        $\mathcal{T}_{\text{hard}}$ & $\mathcal{T}_{\text{glb}}$ & PSNR $\uparrow$ & SSIM $\uparrow$ & LPIPS $\downarrow$ \\  
        \midrule
        \midrule
        \boldcheckmark & - & 20.51 & 0.883 & 0.208 \\
         - & \boldcheckmark & 21.18 & 0.891 & 0.197\\
        \bottomrule
    \end{tabular}
    \vspace{-1em}
    \caption{Impact of the entropy-aware adaptive global threshold on the Patio scene.}
    \label{tab:treshold_glb_mask}
    \vspace{-1.5em}
\end{table}
\begin{figure}
    \centering
    \includegraphics[width=\linewidth]{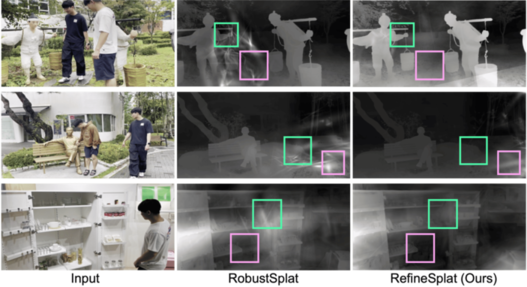}
    \vspace{-2em}
    \caption{Comparison of the depth consistency. We visualize the depth consistency on the Ambiguous wild dataset.}
    \vspace{-1.5em}
    \label{fig:depth_sota}
\end{figure}


%% file: tab_fig/fig_sensitivity.tex
\begin{table}[!t]
    \centering
    \renewcommand{\arraystretch}{0.6}
    \setlength{\tabcolsep}{0.077\linewidth}
    \footnotesize
    \begin{tabular}{cccc}
    \toprule
    $\alpha_1$ & PSNR $\uparrow$ & SSIM $\uparrow$ & LPIPS $\downarrow$   \\
    \midrule
    \midrule
    0.1 & 21.74 & 0.882 & 0.186    \\
    0.2 & 22.11 & 0.897 & 0.173    \\
    0.3 & 21.62 & 0.879 & 0.191    \\
    \bottomrule
    \end{tabular}
    \includegraphics[width=\linewidth]{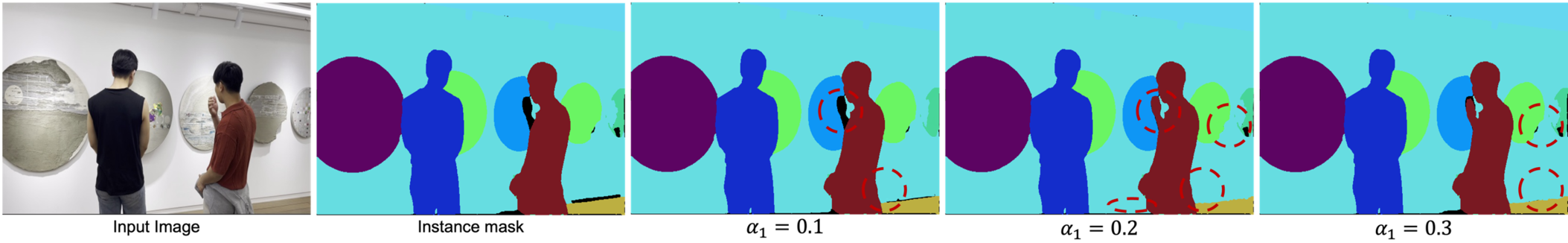}
    \vspace{-2.3em}
    \caption{Sensitivity analysis for $\alpha_1$ values.}
    \label{main_tab:sensitive_cr}
    \vspace{-1.8em}
\end{table}

%% file: tab_fig/fig_abl_regularization_alpha.tex
\begin{figure}[!t]
    \centering
    \includegraphics[width=\linewidth]{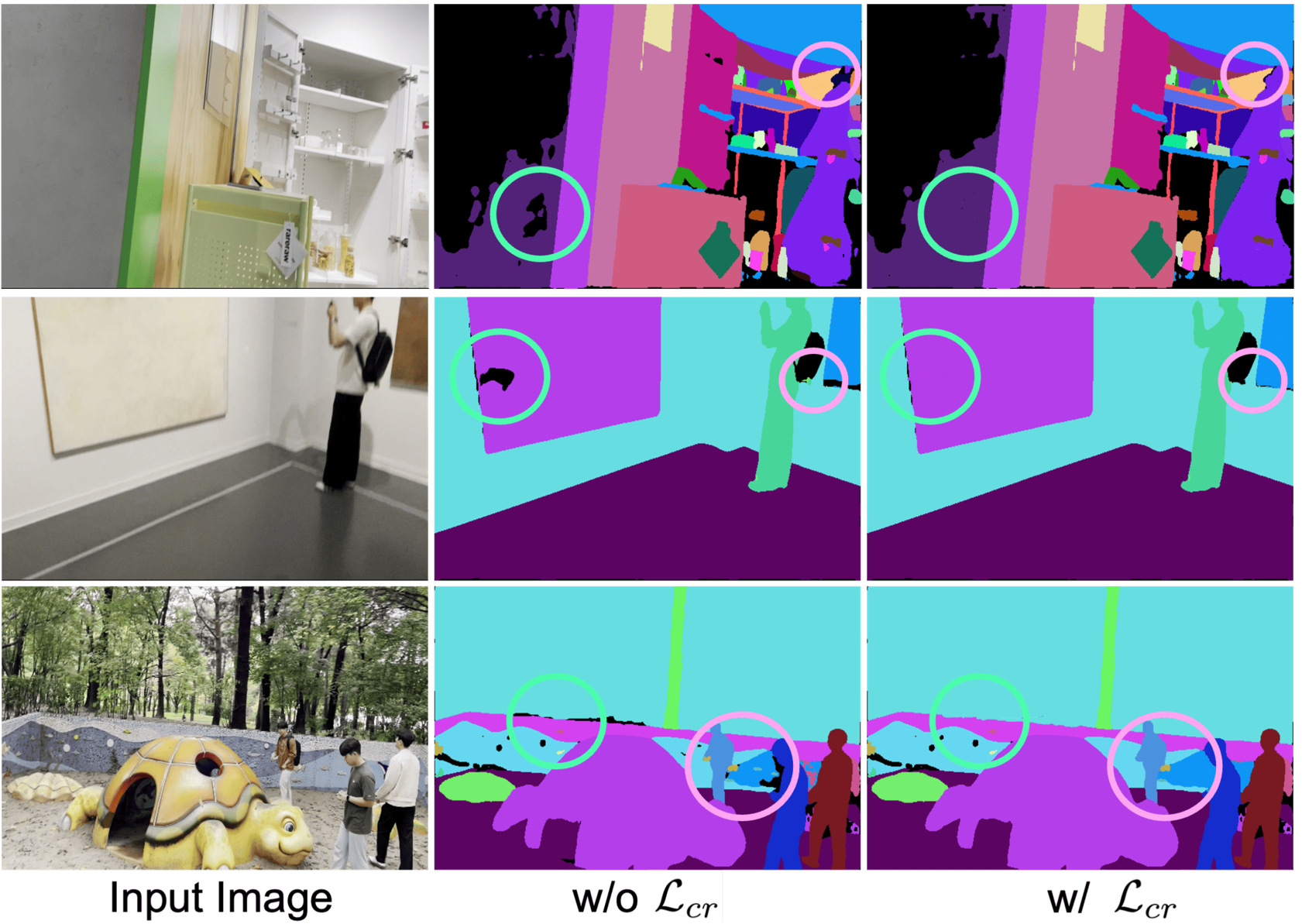}
    \vspace{-2em}    
    \caption{Analysis of the consistency regularization.}
    \label{fig:mask_correction}
    \vspace{-1.2em}
\end{figure}

%% file: tab_fig/tab_feature_sensitivity.tex
\begin{table}[!t]
    \centering
    \resizebox{\linewidth}{!}{
    \begin{tabular}{cccccccccc}
        \toprule
        & \multirow{2}{*}{EAM} & \multirow{2}{*}{EDC} & \multirow{2}{*}{CR} & \multicolumn{3}{c}{Ambiguous wild} & \multicolumn{3}{c}{NeRF On-the-go} \\ 
         &   &  &  & PSNR$\uparrow$ & SSIM$\uparrow$  &LPIPS$\downarrow$ & PSNR$\uparrow$ & SSIM$\uparrow$  & LPIPS$\downarrow$ \\ 
        \midrule
        \midrule
         & \multicolumn{3}{c}{Baseline \cite{fu2025robustsplat}} & 20.52 & 0.756 & 0.277 & 22.14 & 0.746 & 0.166 \\
        (a) & \boldcheckmark & - & - & $22.73$ & $0.834$ &	$0.163$  & $22.78$ & $0.808$ & $0.159$ \\
        (b) & \boldcheckmark & - & \boldcheckmark & $23.46$ & $0.858$ & $0.154$ & $23.13$ & $0.811$ & $0.157$ \\
        (c) & \boldcheckmark & \boldcheckmark & - & $23.14$ & $0.842$ & $0.158$ & $23.29$ & $0.812$ & $0.156$ \\
        (d) & \boldcheckmark  & \boldcheckmark  & \boldcheckmark  & $23.83$ & $0.868$ & $0.149$ & $23.44$ & $0.820$ &	$0.154$ \\
        \bottomrule
    \end{tabular}}
    \includegraphics[width=\linewidth]{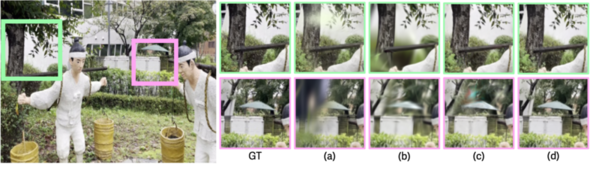} 
    \vspace{-2.2em}
    \caption{Impact of each module. We show impact of each module on NeRF On-the-go and Ambiguous wild datasets.}
    \label{tab:impact_of_module}
    \vspace{-1.3em}
\end{table}

\begin{table}[!t]
\begin{minipage}[t]{0.49\linewidth}
\centering
    \renewcommand{\arraystretch}{0.6}
    \setlength{\tabcolsep}{4pt}
    \tiny
    \begin{tabular}{cccc}
    \toprule
    Methods & PSNR $\uparrow$ & SSIM $\uparrow$ & LPIPS $\downarrow$  \\
    \midrule
    \midrule
    DINO $14 \times 14$  & 23.44 & 0.820 & 0.154 \\
    DINO $16 \times 16$  & 23.37 & 0.817 & 0.149 \\
    \bottomrule
    \end{tabular}
    \vspace{-2em}
    \caption{Sensitivity of size.}
    \label{tab:dino_patch_size}
\end{minipage}
\hfill
\begin{minipage}[t]{0.49\linewidth}
    \centering
    \renewcommand{\arraystretch}{0.6}
    \setlength{\tabcolsep}{4pt}
    \tiny
    \begin{tabular}{cccc}
    \toprule
    Threshold & PSNR $\uparrow$ & SSIM $\uparrow$ & LPIPS $\downarrow$  \\
    \midrule
    \midrule
     DINOv1 & 23.64  & 0.865  &0.131 \\
     DINOv2 & 24.30 & 0.873 & 0.128 \\
     \bottomrule
    \end{tabular}
    \vspace{-2em}
    \caption{Sensitivity of DINO.}
    \label{tab:dino_variant}
\end{minipage}
\vspace{-1em}
\end{table}

%% file: tab_fig/tab_sensitivitiy_parameters.tex
\begin{table}[!t]
\begin{minipage}[t]{0.49\linewidth}
\centering
    \renewcommand{\arraystretch}{0.2}
    \setlength{\tabcolsep}{5pt}
    \tiny
    \begin{tabular}{cccc}
    \toprule
    Filter Size &  PSNR $\uparrow$ &SSIM $\uparrow$ & LPIPS  \\
    \midrule
    \midrule
    3 $\times$ 3 & 21.74 & 0.882 & 0.186    \\
    5 $\times$ 5  & 22.10 & 0.897 & 0.173   \\
    7 $\times$ 7  & 21.32 & 0.853 & 0.192  \\
     \bottomrule
    \end{tabular}
    \vspace{-2em}
    \caption{Sensitivity of filters.}
    \label{tab:dilated_filter}
\end{minipage}
\hfill
\begin{minipage}[t]{0.49\linewidth}
    \centering
    \renewcommand{\arraystretch}{0.2}
    \setlength{\tabcolsep}{4pt}
    \tiny
    \begin{tabular}{cccc}
    \toprule
    Threshold & PSNR $\uparrow$ & SSIM $\uparrow$ & LPIPS $\downarrow$  \\
    \midrule
    \midrule
    0.10 & 23.64 & 0.895 & 0.154 \\
    0.15 & 23.87 & 0.902 & 0.143 \\
    0.20 & 23.45 & 0.891 & 0.157 \\
    \bottomrule
    \end{tabular}
    \vspace{-2em}
    \caption{Threshold of EDC.}
    \label{rebuttal_tab:treshold}
\end{minipage}
\vspace{-1.2em}
\end{table}

%% file: section/conclusion.tex
\input{tab_fig/fig_tab_dataset}
\section{Limitation and Conclusion}
\textbf{Limitations.} RefineSplat captures ambiguous elements leveraging entropy and instance masks. However, there are still two limitations in some cases. Firstly, since RefineSplat leverages vision foundation models, the performance can fluctuate depending on foundation models. Secondly, utilizing sparse views still struggles to generalize visual results. Future work may integrate diffusion models to inpaint sparse or corrupted regions in the Gaussian field.

\vspace{0.2\baselineskip}
\noindent \textbf{Conclusion.} We present RefineSplat, a novel framework for handling ambiguous distractors in diverse real-world scenarios. 
RefineSplat captures ambiguous distractors with Entropy-aware adaptive masking and aligns Gaussians through Entropy-aware density control. We also release the Ambiguous wild dataset with 18 scenes where distractors and static scenes are difficult to distinguish due to color or semantic ambiguity. Extensive results demonstrate the robustness and capability of our method. 

%% file: tab_fig/fig_tab_dataset.tex
\begin{figure}[!t]
    \centering \includegraphics[width=\linewidth]{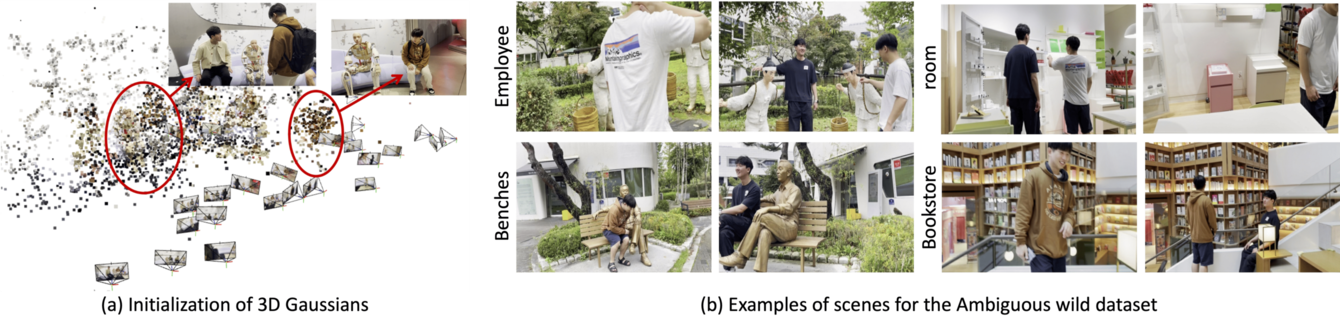}
    \vspace{-2em}
    \caption{Analysis of the Ambiguous wild dataset. (a) Initialization of  Gaussians. (b) Examples of the Ambiguous wild dataset.} 
    \label{fig:data_description}
    \vspace{-1.5em}
\end{figure}

%% file: appendix.tex
\clearpage
\setcounter{page}{1}
\appendix

\maketitle

In this supplementary material, we provide additional discussion and experimental results due to page limitations. Moreover, we also discuss future directions. The contents are summarized as follows:
\begin{itemize}
    \item Sec.~\ref{supple_sec:add_related}: Key distinctions.
    \item Sec.~\ref{supple_sec:dataset}: Descriptions of the Ambiguous wild dataset.
    \item Sec.~\ref{sub:imp_detail}: Implementation details.
    \item Sec.~\ref{sub:exp_detail}: Experiment results.
    \item Sec.~\ref{sub:abl_detail}: Ablation studies.
    \item Sec.~\ref{sub:limit_future}: Future works.
\end{itemize}

\input{supple/add_related}
\input{supple/detail_dataset}
\input{supple/Impl_detail}
\input{supple/exp_detail}

\input{supple/abl_detail}
\input{supple/limit_future}

\newpage

\input{supple_fig/sup_tab_perform}
\input{supple_fig/sup_tab_memory}

\begin{figure*}[!t]
\centering
\includegraphics[width=\textwidth]{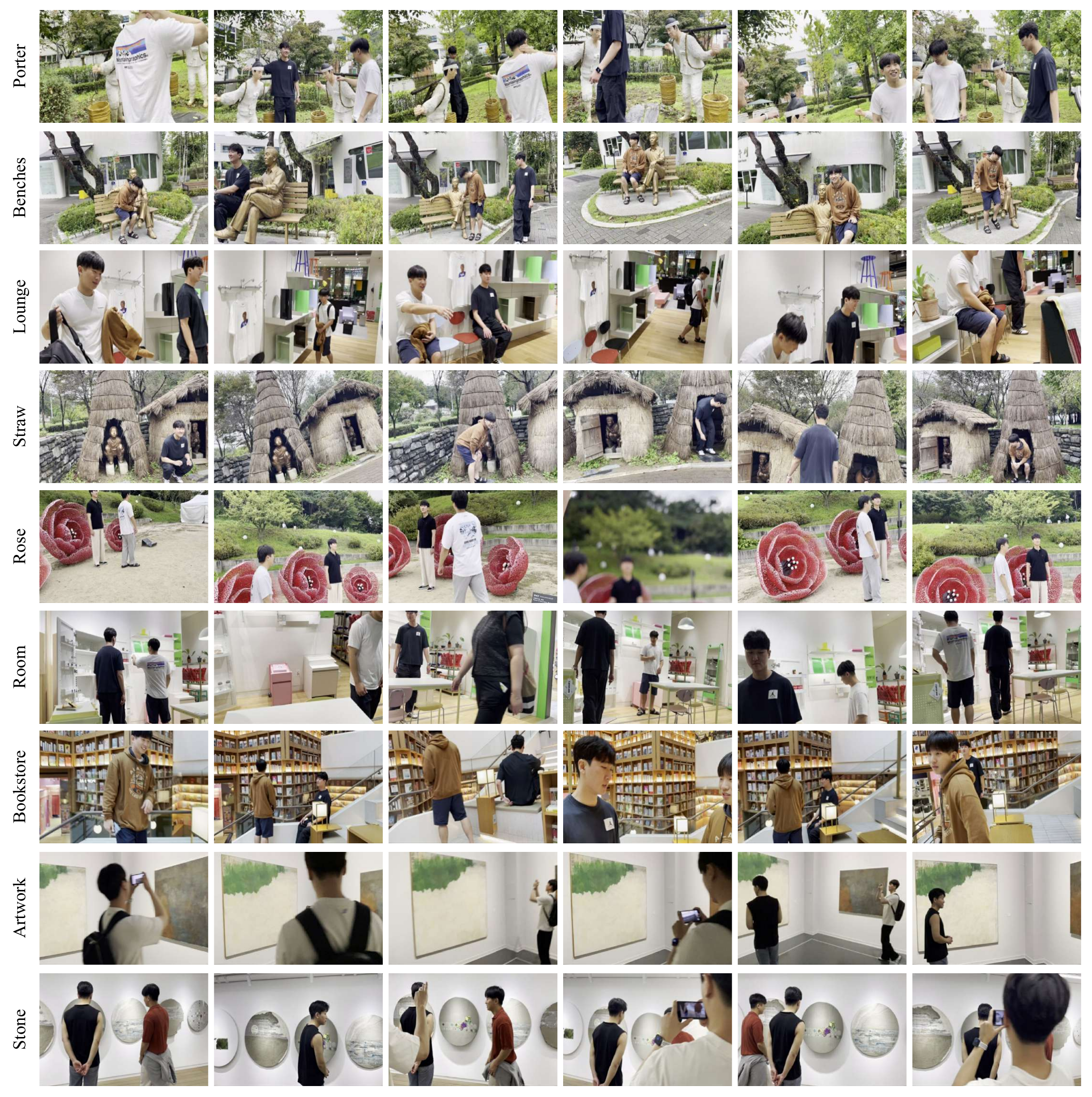}
\caption{Additional sample images on the Ambiguous wild dataset.}
\label{sub_fig:data_sample_1}
\end{figure*}

\begin{figure*}[!t]
\centering
\includegraphics[width=\textwidth]{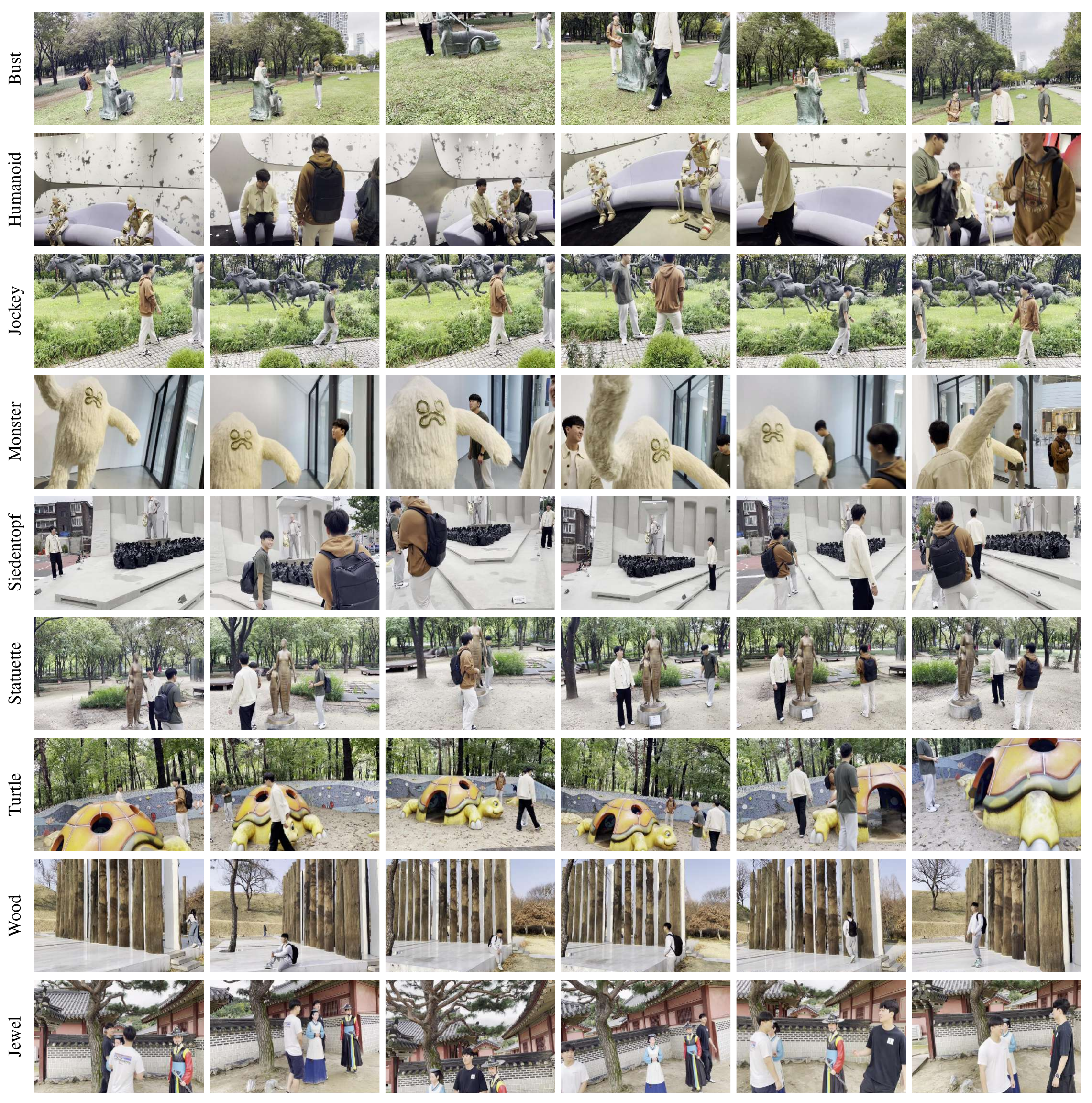}
\caption{Additional sample images on the Ambiguous wild dataset.}
\label{sub_fig:data_sample_2}
\end{figure*}

\begin{figure*}[!t]
\centering
\includegraphics[width=\textwidth]{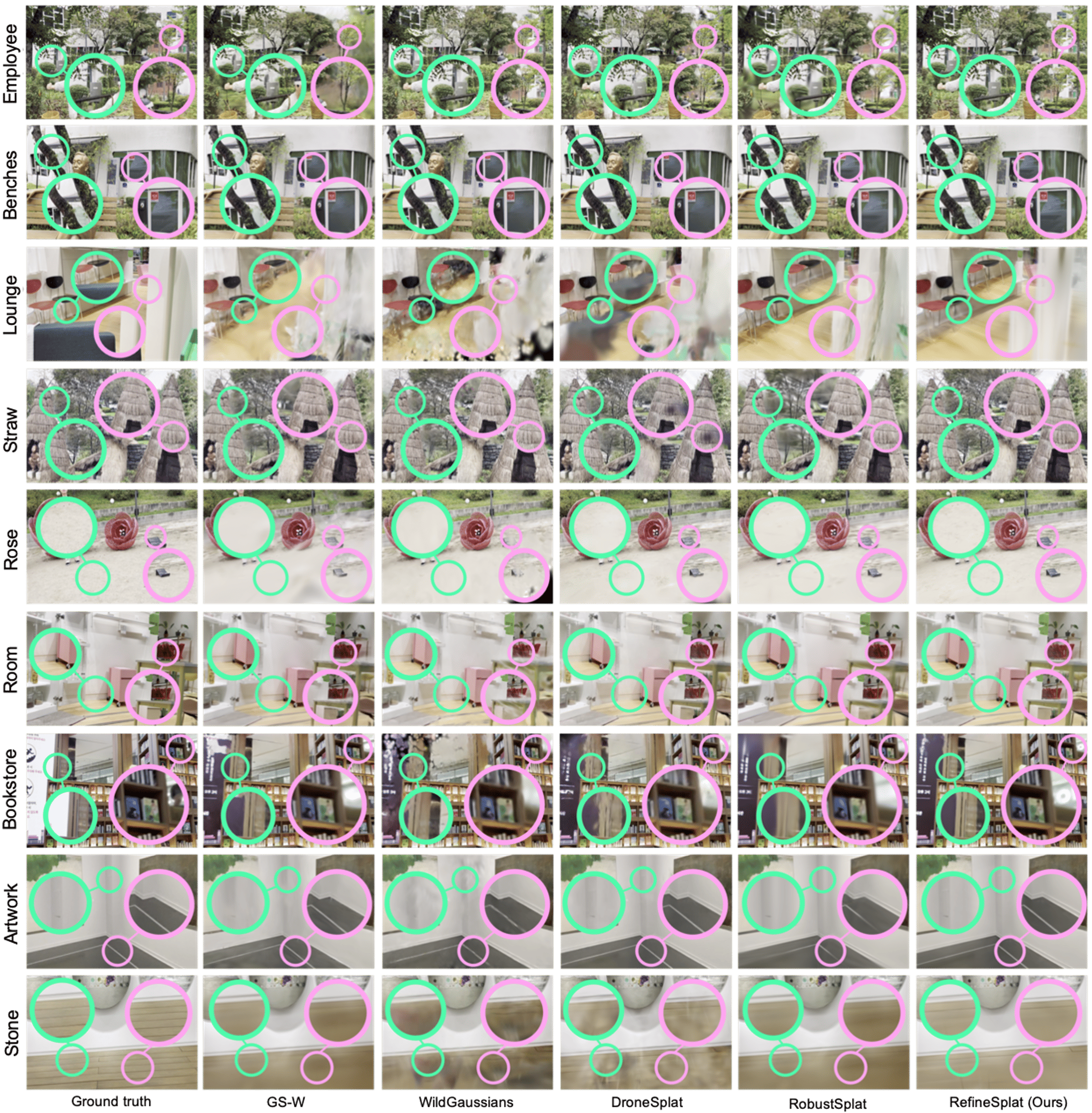}
\caption{Additional qualitative results on the Ambiguous wild dataset.}
\label{sub_fig:ambi_1_qual}
\end{figure*}

\begin{figure*}[!t]
\centering
\includegraphics[width=\textwidth]{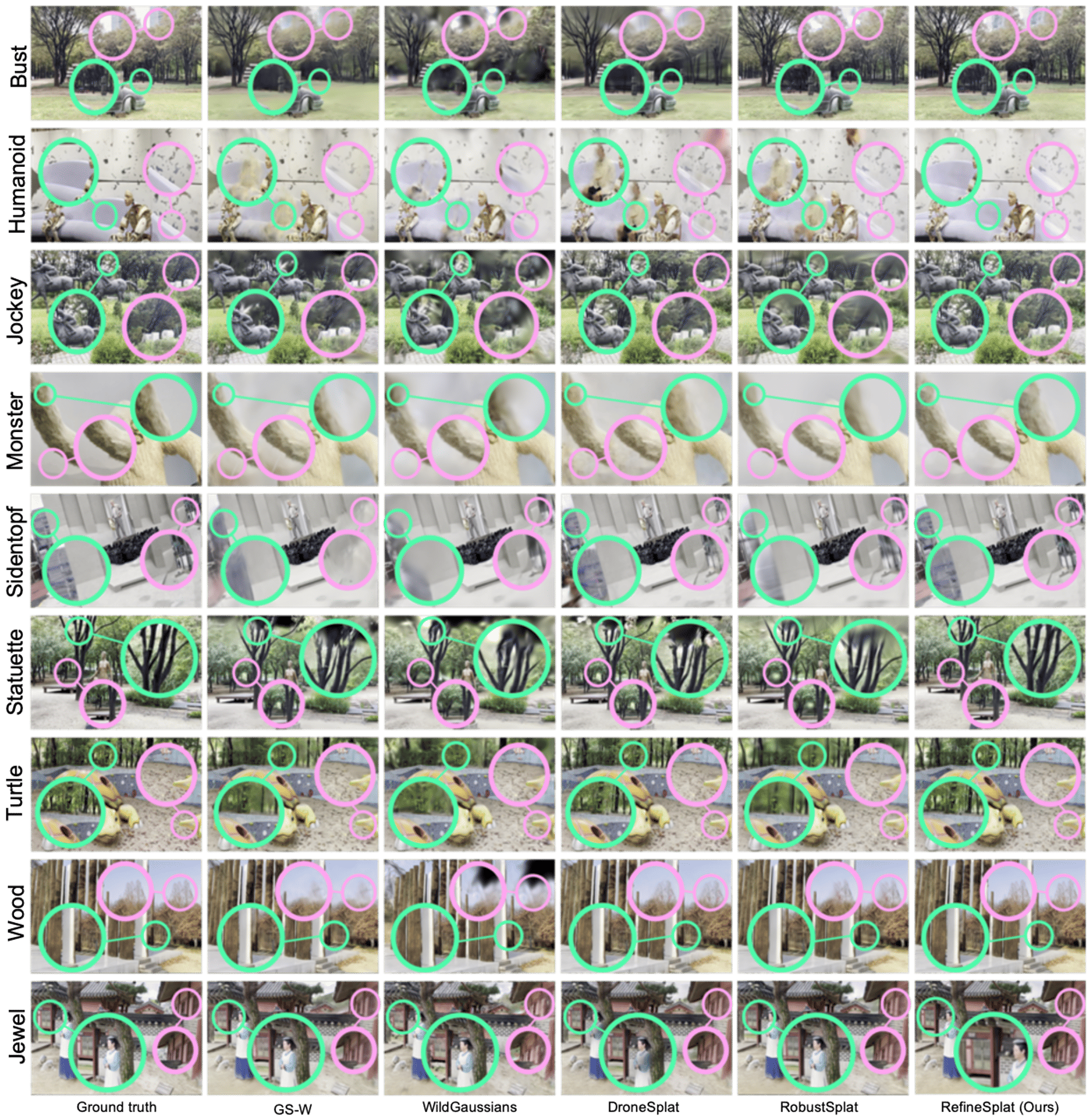}
\caption{Additional qualitative results on the Ambiguous wild dataset.}
\label{sub_fig:ambi_2_qual}
\end{figure*}

\begin{figure*}[!t]
\centering
\includegraphics[width=\textwidth]{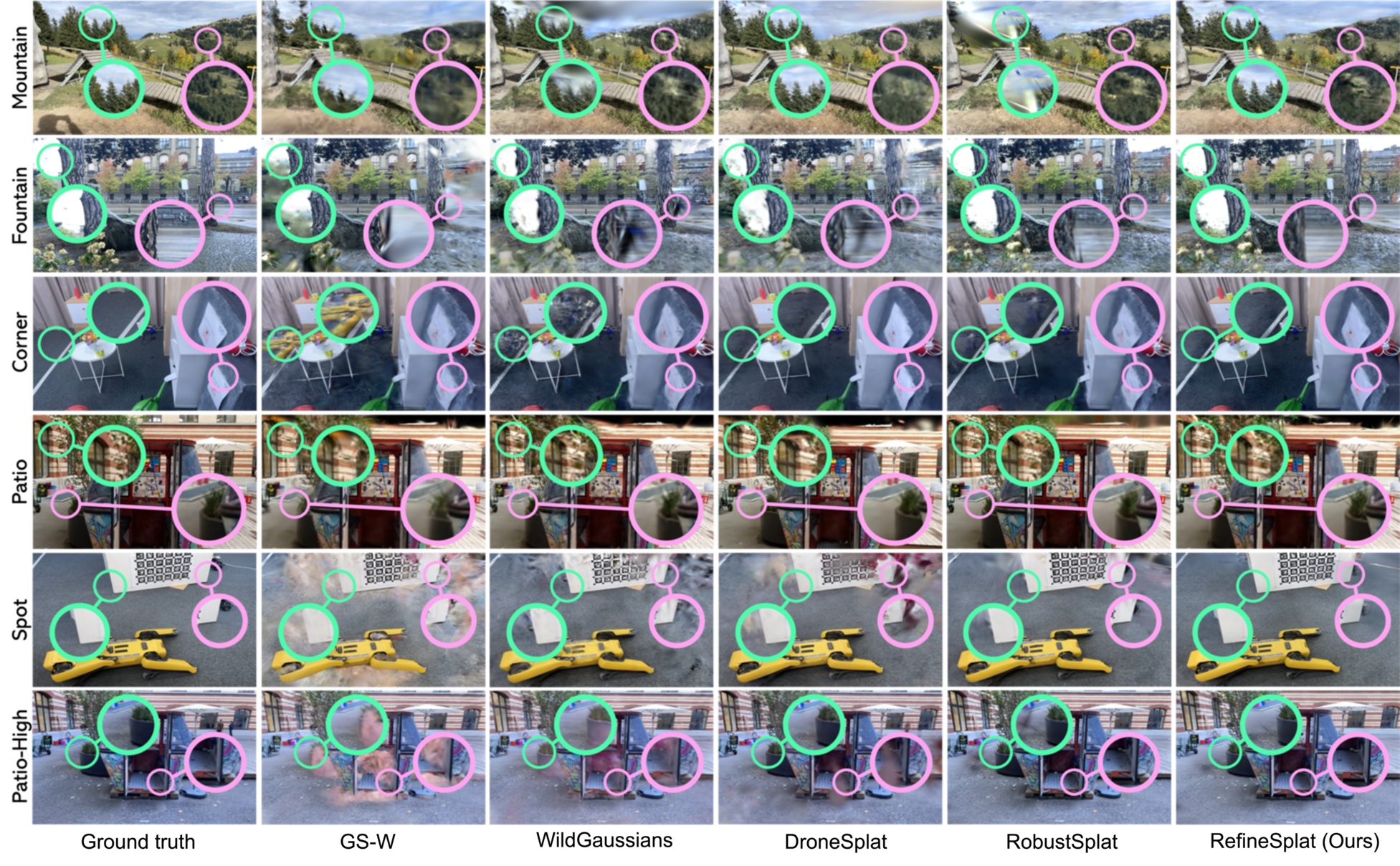}
\caption{Additional qualitative results on the NeRF On-the-go dataset.}
\label{fig:supple_onthego_qual}
\end{figure*}

\begin{figure*}[!t]
\centering
\includegraphics[width=\textwidth]{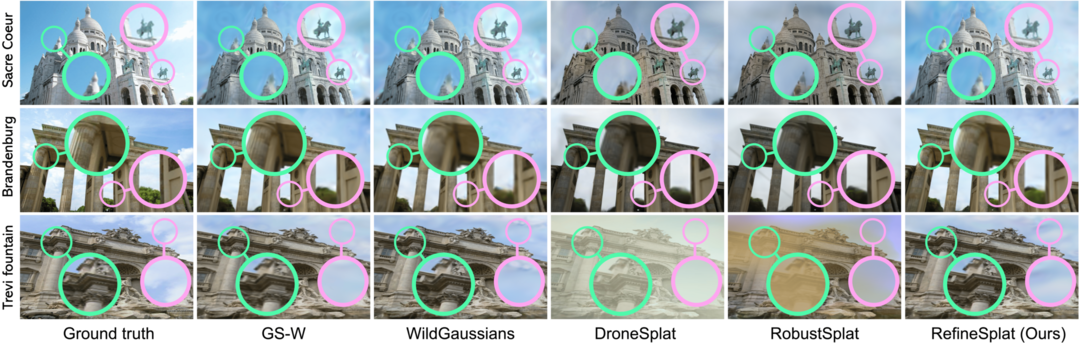}
\caption{Additional qualitative results on the Photo Tourism dataset.}
\label{fig:supple_photo_qual}
\end{figure*}

\begin{figure*}[!t]
\centering
\includegraphics[width=\textwidth]{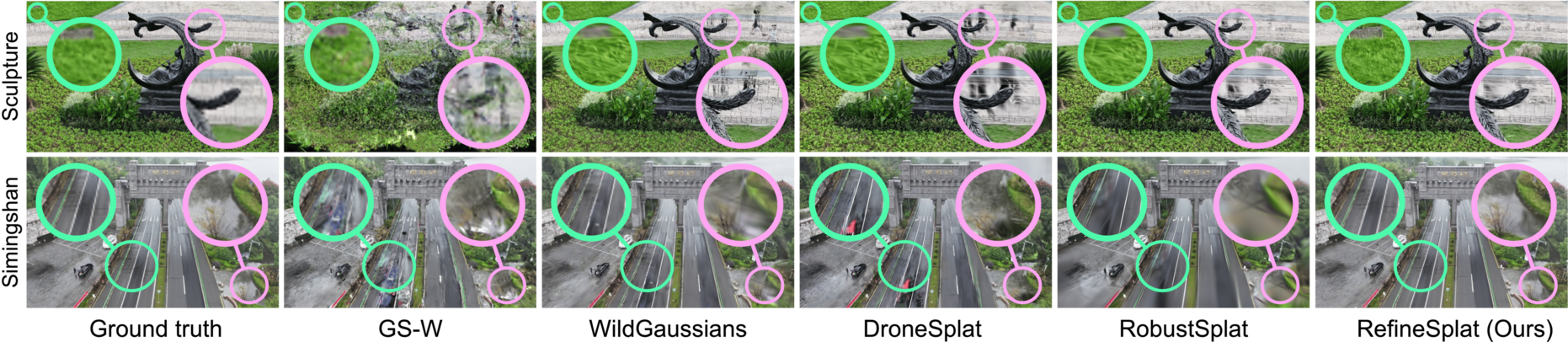}
\caption{Additional qualitative results on the DroneSplat dataset.}
\label{fig:drone_supple_qual}
\end{figure*}

\begin{figure*}[!t]
\centering
\includegraphics[width=\textwidth]{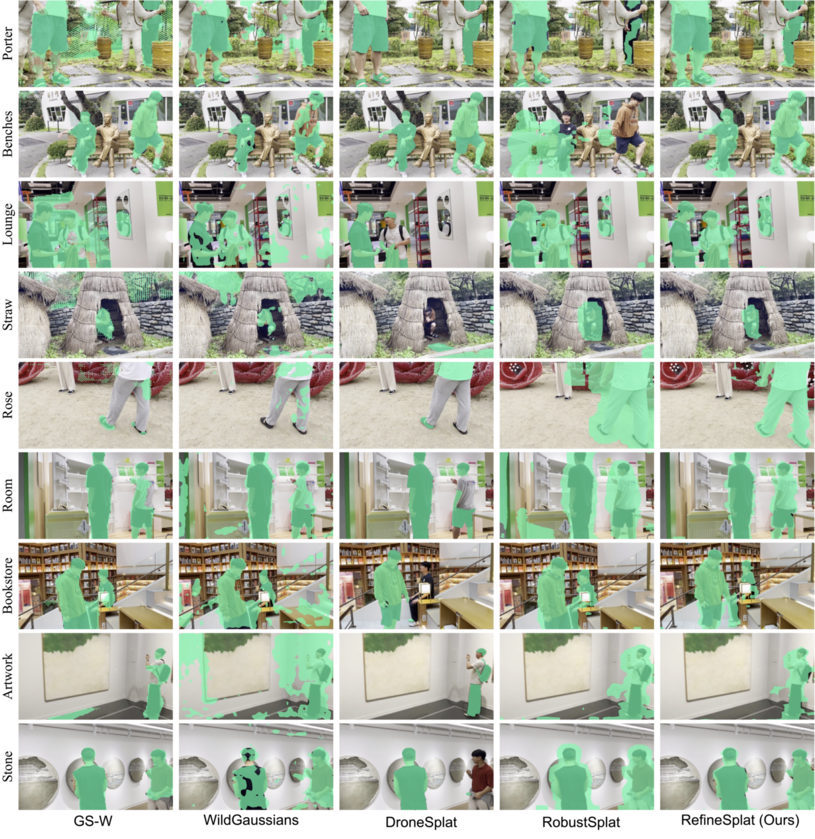}
\caption{Comparison of transient masks in the Ambiguous wild dataset.}
\label{sub_fig:ambi_1_mask}
\end{figure*}

\begin{figure*}[!t]
\centering
\includegraphics[width=\textwidth]{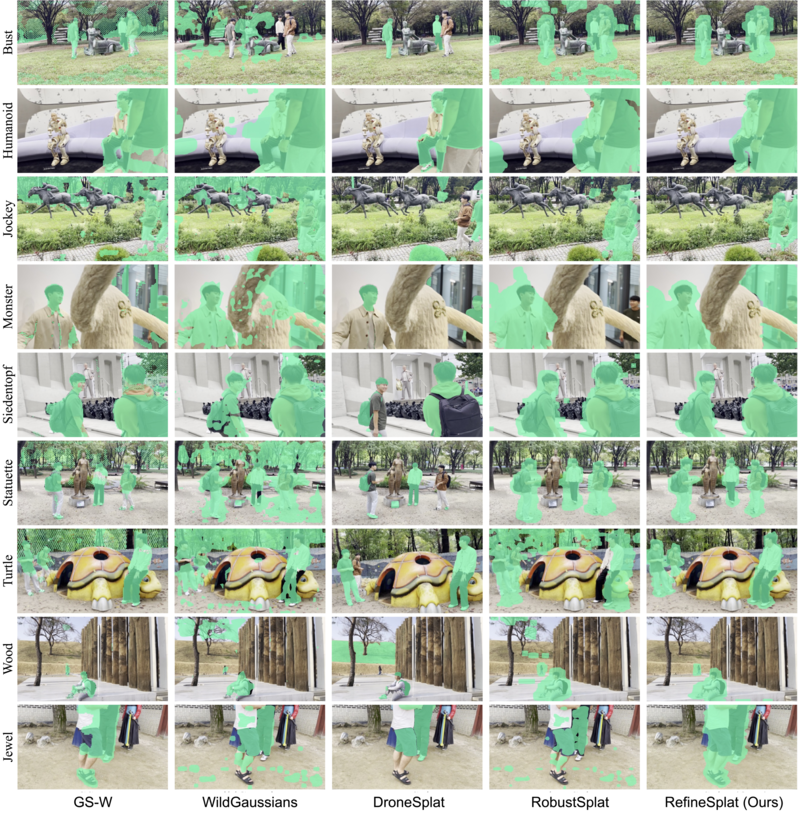}
\caption{Comparison of transient masks in the Ambiguous wild dataset.}
\label{sub_fig:ambi_2_mask}
\end{figure*}

\begin{figure*}[!t]
\centering
\includegraphics[width=\textwidth]{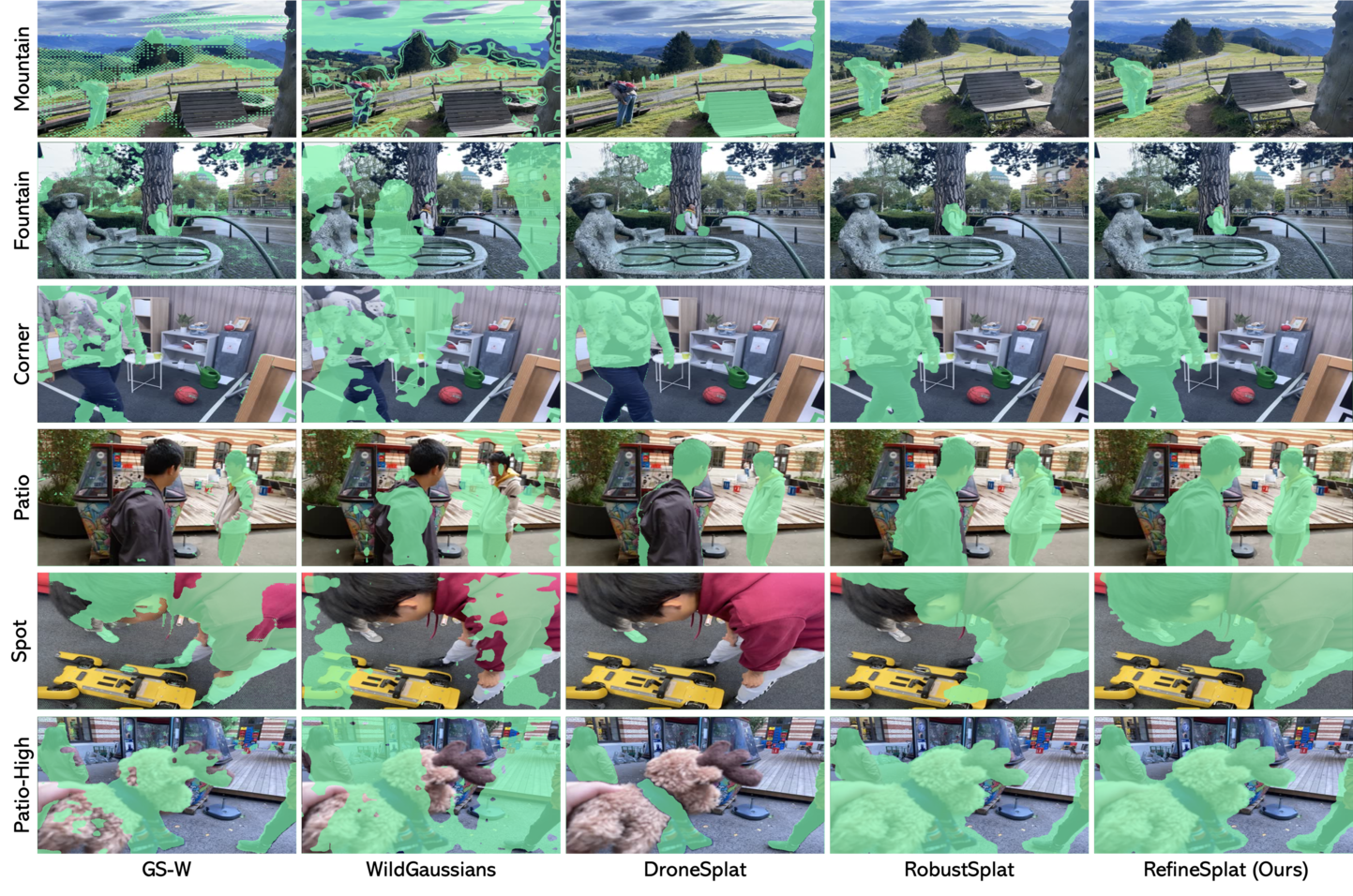}
\caption{Comparison of transient masks in the NeRF On-the-go dataset.}
\label{fig:supple_onthego_Mask}
\end{figure*}

\begin{figure*}[!t]
\centering
\includegraphics[width=\textwidth]{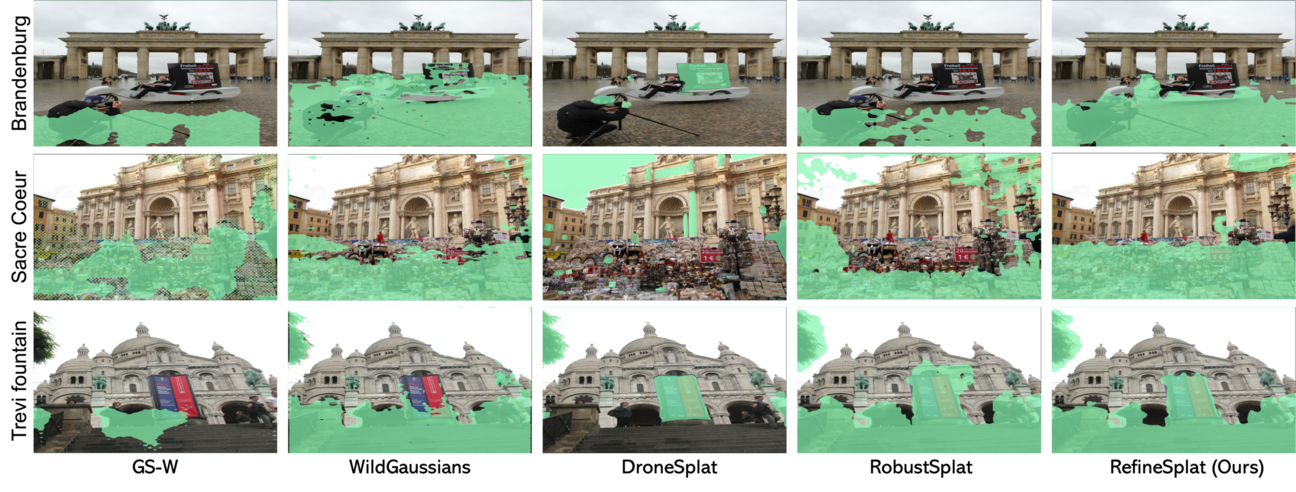}
\caption{Comparison of transient masks in the Photo Tourism dataset.}
\label{fig:supple_photo_mask}
\end{figure*}

\begin{figure*}[!t]
\centering
\includegraphics[width=\textwidth]{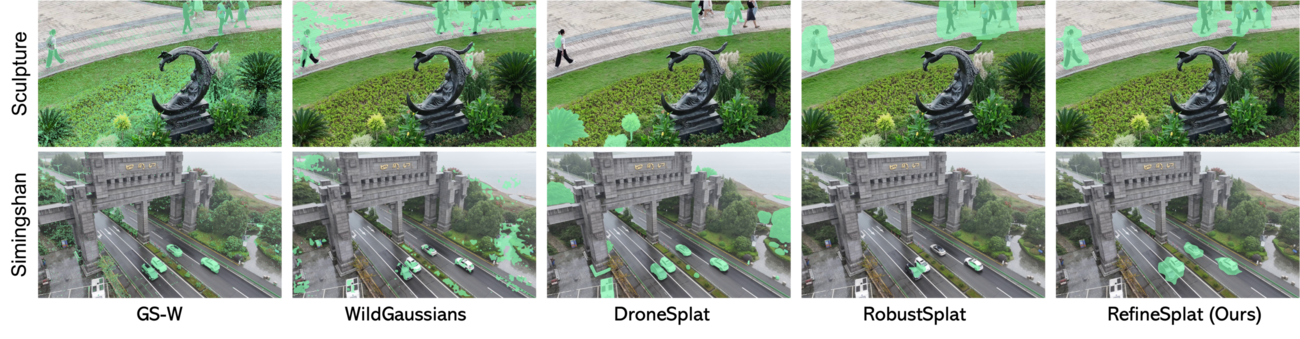}
\caption{Comparison of transient masks in the Drone Imagery dataset.}
\label{fig:drone_supple_mask}
\end{figure*} 

\begin{figure*}[!t]
\centering
\includegraphics[width=\textwidth]{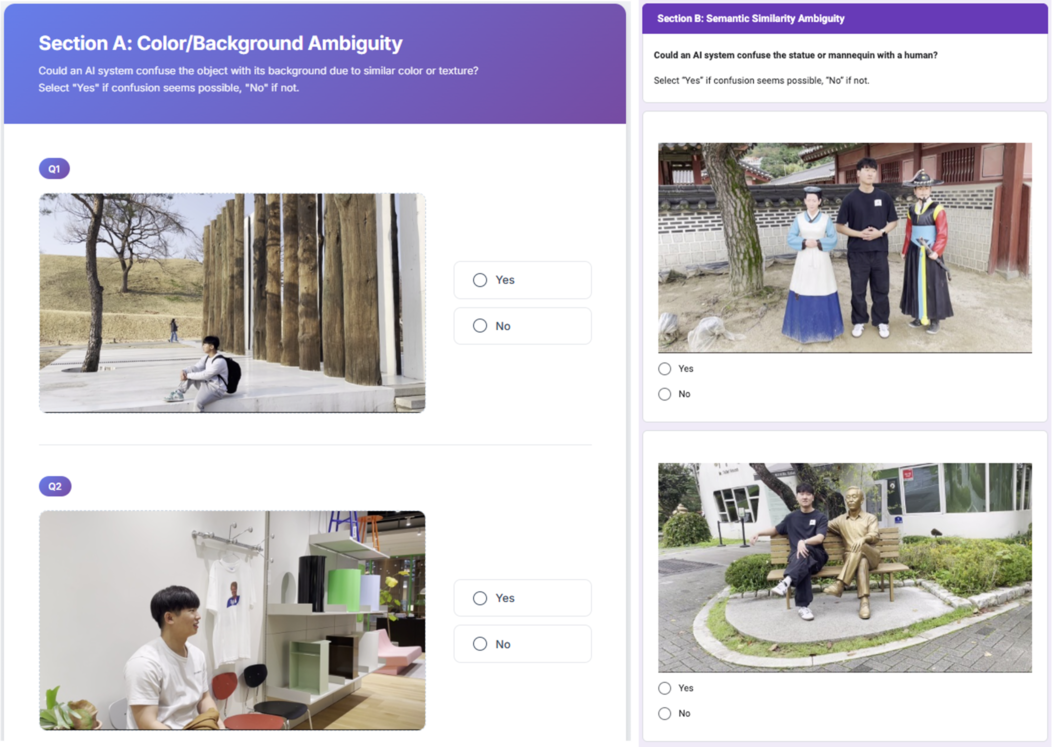}
\vspace{-1.5em}
\caption{Our user study questionnaire. Each participant was shown an upper figure, which is a rendering video of several scenes using different methods. After watching the video, they were asked to answer the question below.}
\label{sub_fig:survey}
\end{figure*}

\clearpage 

\input{supple_fig/sub_tab_notation}





%% file: supple/add_related.tex
\section{Key distinctions}
\label{supple_sec:add_related}

\noindent \textbf{Key distinction from DroneSplat.} Although RefineSplat leverages Grounded SAM \cite{cheng2023tracking}, there are significant technical differences from DroneSplat \cite{tang2025dronesplat}. (1) DroneSplat constructs transient masks using instance masks with photometric error, which makes it hard to capture ambiguous distractors with similar color and semantics to static elements. However, our method effectively captures ambiguous distractors considering entropy. (2) DroneSplat densifies Gaussians considering photometric error, which misaligns and causes artifact issues in the Gaussian field. In contrast, RefineSplat shifts Gaussians considering entropy, which aligns them at the instance level. (3) DroneSplat utilizes statistical measures to construct a threshold in capturing distractors. However, it is hard to identify distractors in a skewed distribution. Compared to DroneSplat and several methods \cite{yang2023cross, ren2024nerf, kulhanek2024wildgaussians} that utilize a fixed threshold, RefineSplat captures ambiguous distractors and consistently shows robustness in diverse real-world scenarios, as shown in Fig. \ref{sub_fig:supple_threshold}.

\vspace{0.2\baselineskip}
\noindent \textbf{Key distinction from SpotlessSplats.} SpotLessSplats \cite{sabour2024spotlesssplats} utilizes semantic features to capture distractors. Despite RefineSplat also leveraging semantic features, it significantly differs from SpotLessSplats in constructing masks. (1) SpotLessSplats highly depends on extracted semantic features from the diffusion model to construct transient masks using photometric error as a condition in training MLPs. However, RefineSplat constructs robust masks by considering instance masks and entropy. (2) SpotLessSplats uses fixed thresholds to define bounds for photometric error, constructing masks with a fixed percentile value. However, RefineSplat adaptively adjusts a threshold considering a skew distribution, showing impressive results in diverse scenarios.

\vspace{0.2\baselineskip}
\noindent \textbf{Key distinction from compression methods.} Several works \cite{huang2025spectral, mallick2024taming, bulo2024revising, lee2024compact} focus on memory compression through hashing and pruning in constrained scenes. RefineSplat densifies and merges Gaussians to effectively align Gaussians, enhancing rendering qualities in real-world scenarios that include diverse distractors. In the process, RefineSplat also achieves memory efficiency as a side effect, showing high-quality rendering results.

\input{supple_fig/sup_fig_threshold}

%% file: supple_fig/sup_fig_threshold.tex
\begin{figure}[!t]
  \centering
  \includegraphics[width=\linewidth]{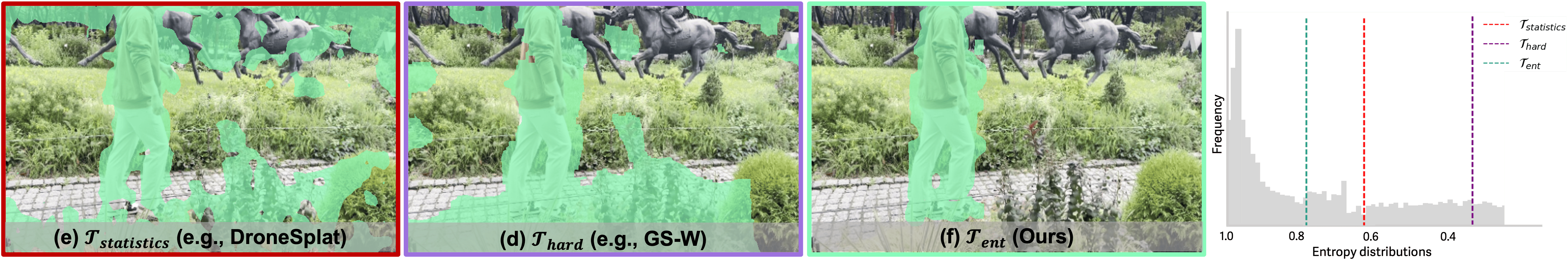}
  \vspace{-1.5em}
  \caption{Analysis of threshold strategies. We visualize how threshold strategies affect the robustness of the transient mask in ambiguous scenarios.}
  \vspace{-1em}
  \label{sub_fig:supple_threshold}
\end{figure}

%% file: supple/detail_dataset.tex
\section{Descriptions of the Ambiguous wild dataset}
\label{supple_sec:dataset}
\noindent \textbf{Details of dataset acquisition.} The Ambiguous wild dataset consists of 18 scenes captured with an iPhone 12 Pro and a Samsung Galaxy S22. The acquisition process follows previous methodologies \cite{sabour2023robustnerf, ren2024nerf} to ensure consistency. During each capture, exposure, white balance, and ISO were kept fixed. This dataset contains diverse visually ambiguous scenarios where transient and static elements exhibit similar colors or semantic levels (e.g., moving vs parked vehicles, a real person vs a mannequin). Furthermore, scenes include varying occlusion ratios ranging from 5\% to 45\%. The images were captured using an iPhone 12 Pro and a Samsung Galaxy S22, with all images recorded at a resolution of 1920×1080. We show samples of each scene in Fig. \ref{sub_fig:data_sample_1} and Fig. \ref{sub_fig:data_sample_2}. 
\input{supple_fig/sup_fig_photo_dino_dataset}

\vspace{0.2\baselineskip}
\noindent \textbf{Evaluation of ambiguity.} To evaluate the ambiguity of the Ambiguous wild dataset, we consider three criteria: residual error, cosine similarity, and human evaluation (User Study). For color similarity, we define color ambiguity as occurring when more than 30\% of the pixels have a color difference of less than 0.3 between distractors and the background, distinguishing transient elements from the static background. For semantic similarity, a scene is considered ambiguous if the cosine similarity between distractors and static objects (e.g.,  sculptures or mannequins) exceeds 70\%, as presented in Fig. \ref{sub_fig:photo_dino_dataset}. Furthermore, to rigorously evaluate ambiguity in our Ambiguous wild dataset, we conduct a user study, as shown in Fig. \ref{sub_fig:survey}. The results from the human evaluation, conducted via a Google Form survey and Amazon Mechanical Turk, clearly demonstrate the presence of ambiguity in our Ambiguous wild dataset, including objects that exhibit similar semantic levels or colors compared to static scenes. Moreover, we observe that leveraging COLMAP \cite{schonberger2016structure} on the Ambiguous wild dataset, where distractors and static elements are visually or semantically similar, tends to produce incorrect initialization of Gaussians, as shown in Fig. \ref{sub_fig:ColMap}.

\noindent \textbf{Author statement of data license.} The authors of the Ambiguous wild dataset affirm that the dataset was collected, processed, and released in full compliance with applicable ethical guidelines and regulations. We assume full responsibility for any potential violations of rights or ethical standards arising from its use. The Ambiguous wild dataset is made freely available under the CC BY-NC-ND 4.0 license, allowing unrestricted sharing and downloading in any medium and repositories, as long as the authors are properly credited, it is used for non-commercial or research purposes, and no derivative works are made.

%% file: supple_fig/sup_fig_photo_dino_dataset.tex
\begin{figure}[!t]
  \centering
  \includegraphics[width=\linewidth]{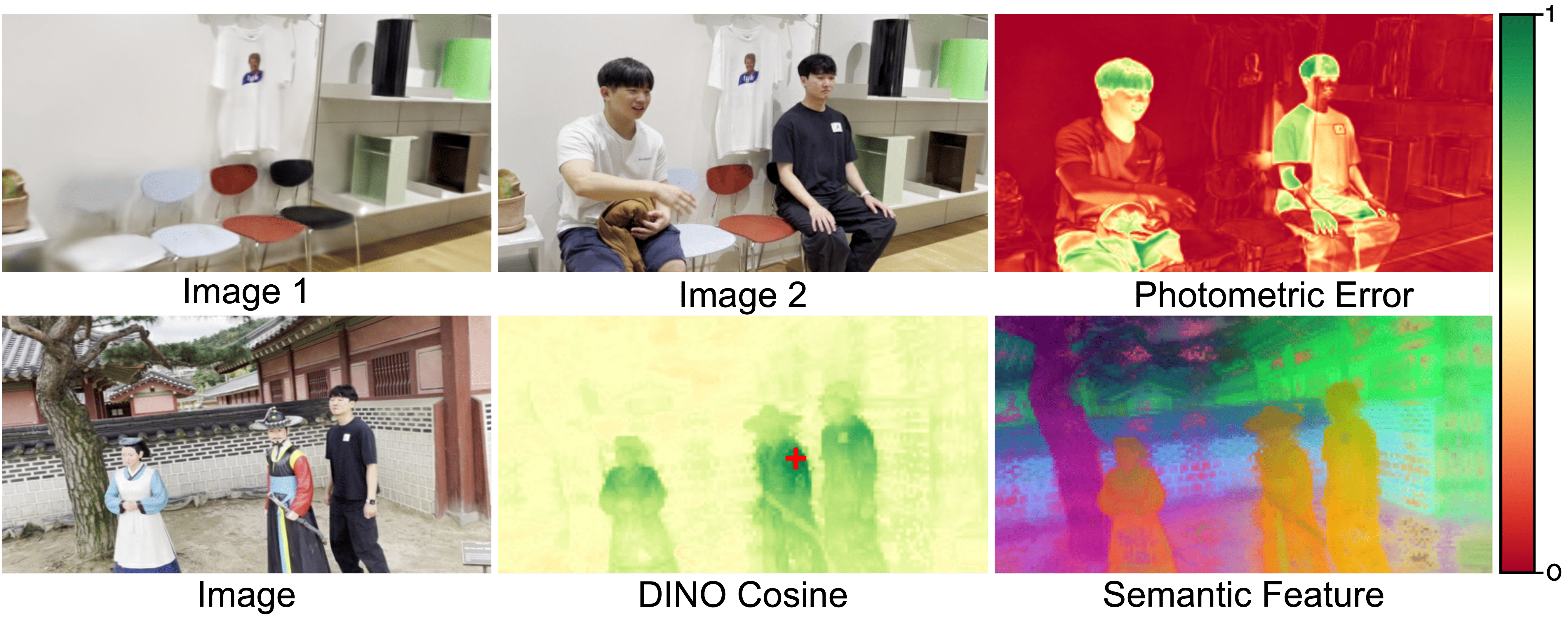}
  \vspace{-2em}
  \caption{Visualization of residual and cosine similarity maps. We color the normalized values, ranging from \BGcolor{5fbaa8}{o}\BGcolor{88cfa4}{n}\BGcolor{b2e0a2}{e}\BGcolor{d7ef9b}{ }\BGcolor{eff8a6}{t}\BGcolor{ffffbf}{o}\BGcolor{feeb9e}{ }\BGcolor{fdd380}{z}\BGcolor{fdb466}{e}\BGcolor{f88d51}{r}\BGcolor{f06744}{o} in ambiguity scenes. The first row is hard to identify static and transients due to color similarity. The last row is challenging to capture transient images due to feature similarity.}
  \vspace{-1em}
  \label{sub_fig:photo_dino_dataset}
\end{figure}

%% file: supple/Impl_detail.tex
\section{Implementation details}
\label{sub:imp_detail}
\textbf{Implementation details.} For training RefineSplat, we conduct the densification interval of 3D Gaussians with hyperparameter settings following RobustSplat \cite{fu2025robustsplat}. For weighting factors $ w_2$, we adopt the sigmoid function to ensure positive constants. To leverage semantic features, we resize the images to extract semantic features from DINOv2 \cite{oquab2023dinov2}. We also use NerfBaselines \cite{kulhanek2024nerfbaselines} for fair comparison. Furthermore, we describe notation to understand our method more easily, as presented in Tab. \ref{sub_tab:notation}. For the frozen MLPs $\psi_{id}$, we pre-train MLPs $\psi_{id}$ using extracted 2D instance masks from Grounded SAM \cite{cheng2023tracking} and semantic features with Adam optimizer \cite{kingma2014adam}, following the main Eq. 10. Moreover, we add a small perturbation to the denominator of main Eq. 6 to prevent the zero-division problem. Moreover, to make understanding easier for readers, we provide procedures of merging in Algorithm \ref{alg:merging}.

\vspace{0.2\baselineskip}
\noindent \textbf{Additional baselines.} We include more baselines (\eg, T-3DGS \cite{markin2024t}, SpotLessSplats \cite{sabour2024spotlesssplats}, DeSplat \cite{wang2024desplat}, HybridGS \cite{lin2024hybridgs}, ForestSplats \cite{park2026forestsplats}, SWAG \cite{dahmani2024swag}) to further validate the effectiveness of our method. Note that for methods without publicly available code, we omit memory usage, and the performance results are reported from the original papers.

\input{supple_fig/sup_fig_colmap}

\vspace{0.2\baselineskip}
\noindent \textbf{Details of adaptive entropy threshold.} To address skewed distributions in ambiguous scenarios, we calculate the adaptive entropy threshold using Shannon entropy \cite{shannon1948mathematical} as:
\begin{equation}
\mathcal{E}(\mathcal{H}) = -\sum_{i = 1}^{N}\mathcal{H}_i\log(\mathcal{H}_i).
\end{equation}By utilizing them, we show robustness in constructing transient masks in a skew distribution compared to existing methods, as shown in Fig. \ref{sub_fig:supple_threshold}. 

%% file: supple_fig/sup_fig_colmap.tex
\begin{figure}[!t]
  \centering
  \includegraphics[width=\linewidth]{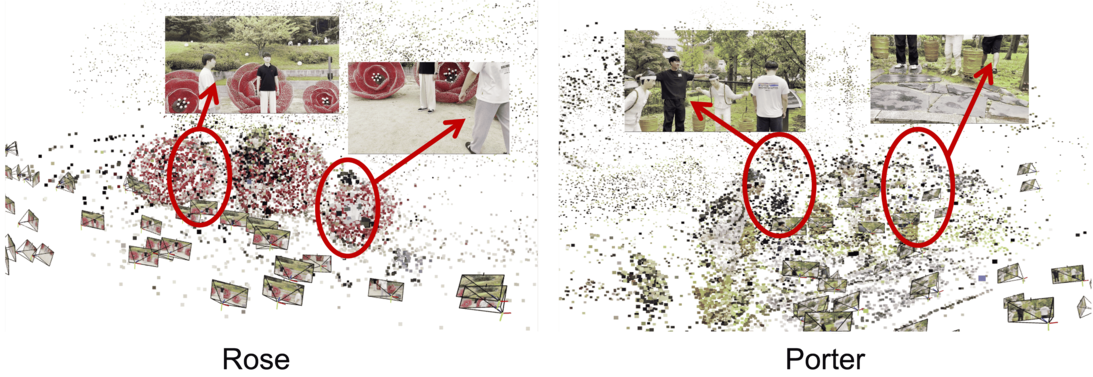}
  \vspace{-2em}
  \caption{Effectiveness of COLMAP SfM point clouds due to ambiguous distractors.}
  \vspace{-1em}
  \label{sub_fig:ColMap}
\end{figure}

%% file: supple/exp_detail.tex
\input{supple_fig/sup_merging_algo}
\section{Experiment results}
\label{sub:exp_detail}
\textbf{Comparison on the Ambiguous wild Dataset.} We present qualitative results and quantitative results, as presented in Fig. \ref{sub_fig:ambi_1_qual}, Fig. \ref{sub_fig:ambi_2_qual}, Tab. \ref{sub_tab:ambi_quan_1}, and Tab. \ref{sub_tab:ambi_quan_2}. Although DroneSplat \cite{tang2025dronesplat} shows reasonable performance, DroneSplat \cite{tang2025dronesplat} has difficulty handling ambiguous scenes where transient and static elements exhibit similar colors or share semantically similar features, due to only considering residual maps to utilize extracted masks from SAM \cite{kirillov2023segment}. Furthermore, DroneSplat \cite{tang2025dronesplat} depends exclusively on extracted transient masks from SAM \cite{kirillov2023segment} before training, making it less capable of handling ambiguous distractors in diverse real-world scenarios. Despite RobustSplat \cite{fu2025robustsplat} and NexusSplats \cite{tang2024nexussplats} showing competitive performances, RobustSplat \cite{fu2025robustsplat} and NexusSplats \cite{tang2024nexussplats} primarily focus on semantic features, which makes it difficult to identify ambiguous distractors that have semantic levels similar to static elements. In contrast, RefineSplat not only effectively removes ambiguous distractors but also shows impressive results leveraging entropy and instance masks. Furthermore, RefineSplat shows effective transient masks in handling ambiguous distractors, as depicted in Fig. \ref{sub_fig:ambi_1_mask} and Fig. \ref{sub_fig:ambi_2_mask}. Furthermore, RefineSplat shows memory efficiency compared to baselines, as shown in Tab. \ref{sub_tab:memory_amb1} and Tab. \ref{sub_tab:memory_amb2}.

\vspace{0.2\baselineskip}
\noindent \textbf{Comparison on the NeRF On-the-go Dataset.} As presented in Fig. \ref{fig:supple_onthego_qual} and Tab. \ref{sub_tab:onthego_quan}, RefineSplat achieves high-quality rendering results and shows competitive performance. Compared to SpotLessSplats \cite{sabour2024spotlesssplats}, which primarily relies on the diffusion process \cite{rombach2022high}, our method exhibits faster training speed while eliminating diverse distractors. Although HybridGS \cite{lin2024hybridgs}, DeSplat \cite{wang2024desplat}, and GS-W \cite{zhang2024gaussian} also show great results, they have difficulty capturing transient elements that possess color similarity. Furthermore, RobustSplat \cite{fu2025robustsplat} and WildGaussians \cite{kulhanek2024wildgaussians} also show impressive results by leveraging semantic features. While semantic features are useful for handling uncertainty in distractors, distinguishing them from static scenes that have a similar semantic level to distractors remains challenging. However, our method leverages instance masks to identify ambiguous distractors, which is effective in capturing diverse distractors, as illustrated in Fig. \ref{fig:supple_onthego_Mask}. Moreover, we show details of memory usage in Tab. \ref{sub_tab:memory_onthego1}. It implies that leveraging the entropy-aware density control improves memory efficiency compared to baselines. Note that we report the total runtime for SpotLessSplats \cite{sabour2024spotlesssplats}, which is the sum of the Diffusion feature extraction time and the training time.

\vspace{0.2\baselineskip}
\noindent \textbf{Comparison on the Photo Tourism Dataset.} We provide additional qualitative and quantitative results, as shown in Fig. \ref{fig:supple_photo_qual} and Tab. \ref{sub_tab:photo_quan}. Although WildGaussians \cite{kulhanek2024wildgaussians} and NexusSplats \cite{tang2024nexussplats} provide quantitatively remarkable results, those methods only focus on interpolated semantic features, which makes it difficult to capture fine-grained details of distractors. While RobustSplat \cite{fu2025robustsplat} and DroneSplat \cite{tang2025dronesplat} also show impressive results on the NeRF On-the-go \cite{ren2024nerf} and Ambiguous wild datasets, they are hard to tackle appearance variation on the Photo Tourism \cite{snavely2006photo} dataset due to a lack of appearance modeling. GS-W \cite{zhang2024gaussian}, SWAG \cite{dahmani2024swag}, and DeSplat \cite{wang2024desplat} also show impressive results by considering only the residual map to identify transient elements. However, it is difficult to detect distractors that have similar colors to static scenes. To understand these challenges, we visualize how the transient mask works, as illustrated in Fig. \ref{fig:supple_photo_mask}. Specifically, our Entropy-aware adaptive masking can capture more fine-grained details compared to prior methods. Moreover, we provide details of memory, as shown in Tab. \ref{sub_tab:memory_photo_drone_1}.

\input{supple_fig/sup_fig_sparse_issue}
\input{supple_fig/sup_fig_appearance}
\vspace{0.2\baselineskip}
\noindent \textbf{Comparison on the Drone Imagery Dataset.} We provide further qualitative and quantitative results, as shown in Fig. \ref{fig:drone_supple_qual} and Tab. \ref{sub_tab:supple_Drone_quan}. While DroneSplat \cite{tang2025dronesplat} shows impressive rendering quality, it struggles to capture ambiguous color similarities between static elements and distractors, as it relies on residual errors to generate invariant masks extracted from SAM \cite{kirillov2023segment} before training. Although RobustSplat \cite{fu2025robustsplat} also shows great results, it only depends on semantic features to capture transient elements, which makes it hard to tackle similar semantic levels to static elements. In contrast, RefineSplat considers distractors at the instance level to better identify ambiguous distractors between transient and static elements. Moreover, we show the efficacy of our entropy-aware adaptive mask on the Drone Imagery dataset \cite{tang2025dronesplat}, as shown in Fig. \ref{fig:drone_supple_mask}.

%% file: supple_fig/sup_merging_algo.tex
\begin{algorithm}[t]
    \KwIn{Rendered image $\hat{I}$, GT image $I^{gt}$, transient masks $\mathcal{M}_{final}$, and instance masks $\mathcal{M}_{ins}$ }
    \textbf{Initialization:} $ \mu, S, R, \alpha, c \gets \text{Structure-from-Motion (SfM)}$\cite{schonberger2016structure} \\
    \For {$i$ \textbf{in} ($1,\ldots,N$)} 
        {
            \tcc{Render and Optimize the 3D representation}
            $\mathcal{L}_{\text{total}} \gets \mathcal{L}_{\text{GS}}(\hat{I}, \tilde{I}^{gt}, \mathcal{M}_{final}) + \mathcal{L}_{\text{id}}(\hat{I}, \tilde{I}^{gt}, \mathcal{M}_{ins}) + \mathcal{L}_{\text{cr}}(\hat{I}, \mathcal{M}_{ins})$  \\
            $\mu, S, R, \alpha, c \gets Adam(\nabla \mathcal{L}_{\text{total}})$ \\
            \If{$\text{IsEntropyDensityControl}(i)$}{
                $\text{Densification}(\mathcal{G})$ \\
                \If{$\text{IsMerging}(i)$ }{
                \For {Gaussians $\mathcal{G}_i(\mu, S, R, \alpha, c)$} {
                    \If{${\nabla \mathcal{L}_{\text{total}}} < \tau_{e}$ } {
                    \For {k-neighbor Gaussians $\mathcal{G}_i(\mu, S, R, \alpha, c)$ \Comment{$i \ne j$}} {
                    $\boldsymbol{O} \gets \text{Overlap Ratio}(\mathcal{G}_i, \mathcal{G}_j)$ \Comment{Main Eq. 11} \\
                        \If{$\text{ColorResidual}(c_i, c_j) < 0.15$ \textbf{and} $\boldsymbol{O} > 0.3$} {
                            $\text{Merging}(\mathcal{G}_i, \mathcal{G}_j)$ \\
                            }
                        }
                    }
                }
                }
            }
        }
    \caption{Merging Pipeline}
    \label{alg:merging}
\end{algorithm}

%% file: supple_fig/sup_fig_sparse_issue.tex
\begin{figure}[!t]
  \centering
  \includegraphics[width=\linewidth]{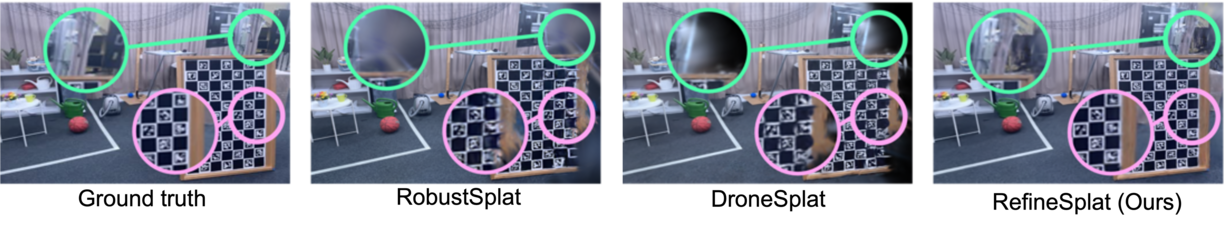}
  \vspace{-2em}
  \caption{An analysis of sparsity issue on the Corner scene.}
  \vspace{-2em}
  \label{sub_fig:sparsity}
\end{figure}

%% file: supple_fig/sup_fig_appearance.tex
\begin{figure*}[!t]
  \centering
  \includegraphics[width=\linewidth]{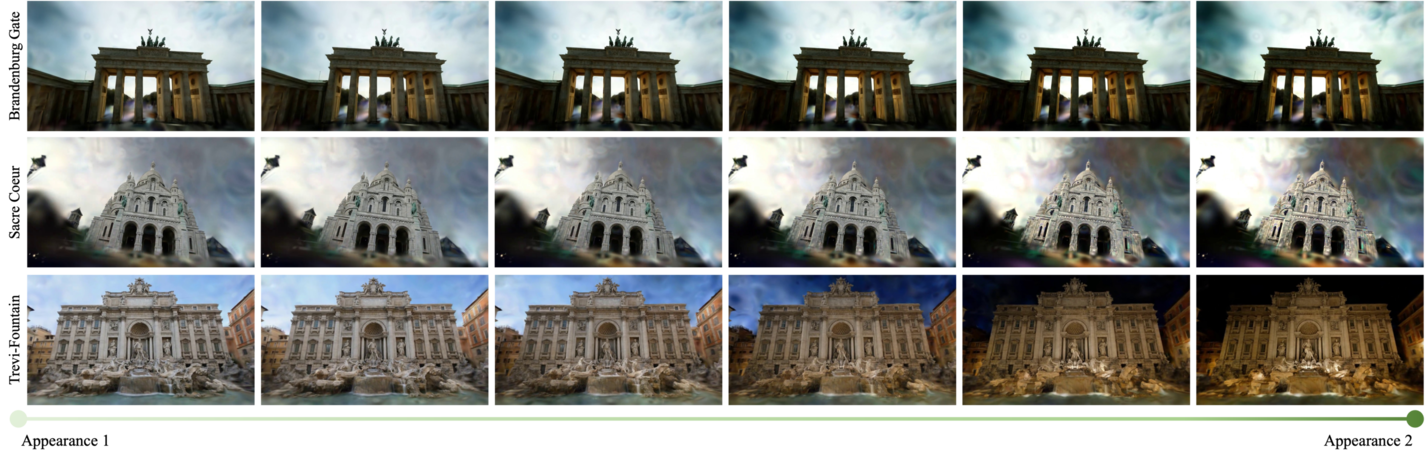}
  \vspace{-2em}
  \caption{Visualization of Appearance variations on the Photo Tourism dataset.}
  \label{sub_fig:appearance}
\end{figure*}

%% file: supple/abl_detail.tex
\section{Ablation studies}
\label{sub:abl_detail}
\noindent \textbf{Robustness of adaptive entropy threshold.} To demonstrate the robustness of the adaptive entropy threshold for variation of the entropy compared to baselines, we show step-by-step proof by differentiating $\mathcal{T}_{ent}$. The derivation for $\mathcal{T}_{ent}$ can be calculated as:
\begin{equation}
\frac{\partial \mathcal{T}_{ent}}{\partial \mathcal{H}_i} \propto- w_2\log \left(\mathcal{H}_i\right).
\label{sub_math:temp}
\end{equation}
On the other hand, the derivation for $\mathcal{T}_{statistics}$, which is utilized in DroneSplat \cite{tang2025dronesplat}, is calculated as follows:
\begin{equation}
\frac{\partial \mathcal{T}_{\text {statistics }}}{\partial \mathcal{H}_i} \propto \frac{2}{N}\left(\mathcal{H}_i-\mu_{\mathcal{H}}\right).
\end{equation} Thus, compared to the derivation for $\mathcal{T}_{statistics}$, utilizing a logarithm to construct transient masks for capturing distractors, $\mathcal{T}_{ent}$ is less sensitive to variation of a skew entropy distribution. Furthermore, we also observe that leveraging $\mathcal{T}_{ent}$ constructs robust transient masks compared to baselines, as shown in Fig. \ref{sub_fig:supple_threshold}. Note that, to focus on the core part of the derivation, we discard constant terms that emerge during the derivation.

\vspace{0.2\baselineskip}
\noindent \textbf{Appearance variations on the Photo Tourism dataset.} Furthermore, we examine whether RefineSplat addresses appearance variations and consistently maintains appearance, as illustrated in Fig. \ref{sub_fig:appearance}. To tackle appearance variations, we follow the appearance modeling strategy from WildGaussians\cite{kulhanek2024wildgaussians}, which utilizes
per-view image embeddings to handle appearance variations with Adam optimizer \cite{kingma2014adam}. The per-view image embeddings are of size 32. For the appearance MLP, we use 2
layers of size 128 with ReLU activations.

%% file: supple/limit_future.tex
\section{Future works}
\label{sub:limit_future}
\noindent \textbf{Future works.} RefineSplat achieves superior results in capturing ambiguous distractors across diverse real-world datasets by leveraging entropy and instance masks. However, similar to existing approaches, our method struggles to generalize in 3D novel-view synthesis from unconstrained sparse training views in diverse real-world scenarios, as shown in Fig. \ref{sub_fig:sparsity}. To address these issues, a feasible solution is to integrate our methods with diffusion models \cite{rombach2022high} to interpolate the sparsity region in the Gaussian field and distill a 3D representation. Moreover, while we demonstrate the effectiveness of RefineSplat on the Photo Tourism \cite{snavely2006photo}, Drone Imagery \cite{tang2025dronesplat}, NeRF On-the-go \cite{ren2024nerf}, and Ambiguous wild datasets, another promising direction is extending RefineSplat to large-scale dynamic scenes such as KITTI360~\cite{liao2022kitti}, NuScenes~\cite{caesar2020nuscenes}, and Cosmos~\cite{agarwal2025cosmos}. Furthermore, since real-world scenarios often contain both blurring issues and diverse distractors, constructing such datasets would be a promising future direction to tackle extreme real-world scenarios. We leave this as future extension works. 

%% file: supple_fig/sup_tab_perform.tex
\begin{table*}[!t]
    \centering
    \setlength{\tabcolsep}{0.1pt}
    \renewcommand{\arraystretch}{1.05}
    \resizebox{\linewidth}{!}{
    \begin{tabular}{lcccccccccccccccccccccccccccc}
    \toprule
    \multirow{2}{*}{Method} & \multirow{2}{*}{GPU hrs / FPS} & \multicolumn{3}{c}{Artwork} & \multicolumn{3}{c}{Benches} & \multicolumn{3}{c}{Bookstore} & \multicolumn{3}{c}{Jewel} & \multicolumn{3}{c}{Monster} & \multicolumn{3}{c}{Wood}\\
                               &&  PSNR$\uparrow$ &SSIM$\uparrow$ & LPIPS$\downarrow$ & PSNR$\uparrow$ &SSIM$\uparrow$ & LPIPS$\downarrow$ & PSNR$\uparrow$ &SSIM$\uparrow$ & LPIPS$\downarrow$ & PSNR$\uparrow$ &SSIM$\uparrow$ & LPIPS$\downarrow$ & PSNR$\uparrow$ &SSIM$\uparrow$ & LPIPS$\downarrow$ & PSNR$\uparrow$ &SSIM$\uparrow$ & LPIPS$\downarrow$  \\
    \midrule
    \midrule
    3DGS \cite{kerbl20233d}& 1.8 / 57 & 27.13 & 0.943 & 0.202 & \metrictablethird{26.64} & \metrictablethird{0.903} & 0.113 & 20.66 & 0.688 & 0.251 & 25.62 & \metrictablethird{0.899} & 0.105 & 20.43 & 0.856 & 0.136 & 17.14 & 0.766 & 0.369 \\
    Mip-Splatting \cite{yu2024mip}& 1.75 / 66 & 28.42 & 0.954 & 0.185 & \metrictablesecond{27.65} & \metrictablesecond{0.913} & \metrictablesecond{0.102} & \metrictablethird{22.24} & \metrictablesecond{0.734} & \metrictablethird{0.231} & 26.07 & \metrictablesecond{0.905} & \metrictablethird{0.099} & 20.37 & 0.872 & 0.129 & 17.88 & 0.795 & 0.344 \\
    GS-W \cite{zhang2024gaussian}& 1.01 / 106 & 31.15 & \metrictablesecond{0.963} & \metrictablethird{0.151} & 25.67 & 0.837 & 0.123 & 20.40 & 0.552 & 0.461 & 26.68 & 0.882 & 0.106 & \metrictablesecond{21.14} & \metrictablesecond{0.893} & \metrictablefirst{0.085} & 18.99 & \metrictablesecond{0.889} & 0.295 \\
    WildGaussians \cite{kulhanek2024wildgaussians}& 1.63 / 102 & 27.22 & 0.943 & 0.197 & 23.68 & 0.794 & 0.193 & 19.59 & 0.457 & 0.585 & 25.21 & 0.849 & 0.152 & 20.91 & 0.855 & 0.147 & 18.42 & 0.844 & 0.326 \\
    NexusSplats \cite{tang2024nexussplats} & 1.31 / 98 & 24.58 & 0.933 & 0.241 & 23.08 & 0.789 & 0.218 & 19.81 & 0.509 & 0.474 & 25.88 & 0.862 & 0.146 & 21.03 & 0.866 & 0.142 & 18.35 & 0.794 & 0.415 \\
    DroneSplat \cite{tang2025dronesplat}& 0.71 / 104 & \metrictablethird{31.44} & 0.958 & \metrictablefirst{0.032} & 26.14 & 0.869 & \metrictablefirst{0.083} & \metrictablesecond{22.40} & \metrictablethird{0.695} & \metrictablefirst{0.202} & \metrictablethird{26.69} & 0.889 & \metrictablefirst{0.066} & 20.68 & 0.861 & \metrictablethird{0.094} & \metrictablethird{19.72} & 0.818 & \metrictablesecond{0.203} \\
    RobustSplat \cite{fu2025robustsplat}& 0.41 / 106 & \metrictablesecond{31.79} & \metrictablethird{0.961} & 0.207 & 25.37 & 0.83 & 0.234 & 21.20 & 0.529 & 0.472 & \metrictablesecond{26.78} & 0.885 & 0.154 & \metrictablethird{21.78} & \metrictablethird{0.890} & 0.119 & \metrictablesecond{21.20} & \metrictablethird{0.887} & \metrictablethird{0.236} \\
    RefineSplat (Ours) & 0.54 / 105 & \metrictablefirst{32.45} & \metrictablefirst{0.975} & \metrictablesecond{0.138} & \metrictablefirst{28.06} & \metrictablefirst{0.915} & \metrictablethird{0.106} & \metrictablefirst{23.24} & \metrictablefirst{0.743} & \metrictablesecond{0.228} & \metrictablefirst{27.88} & \metrictablefirst{0.919} & \metrictablesecond{0.089} & \metrictablefirst{22.20} & \metrictablefirst{0.911} & \metrictablesecond{0.086} & \metrictablefirst{22.68} & \metrictablefirst{0.902} & \metrictablefirst{0.197} \\
    \bottomrule
    \end{tabular}}
    \caption{Quantitative result on the Ambiguous wild dataset. Performance is highlighted by color from \metrictablethird{third} \metrictablesecond{to} \metrictablefirst{first}.}
    \label{sub_tab:ambi_quan_1}
    \vspace{-0.1em}
\end{table*}
\begin{table*}[!t]
    \centering
    \setlength{\tabcolsep}{0.1pt}
    \renewcommand{\arraystretch}{1.05}
    \resizebox{\linewidth}{!}{
    \begin{tabular}{lcccccccccccccccccccccccccccc}
    \toprule
    \multirow{2}{*}{Method} & \multirow{2}{*}{GPU hrs / FPS} & \multicolumn{3}{c}{Room} & \multicolumn{3}{c}{Rose} & \multicolumn{3}{c}{Turtle} & \multicolumn{3}{c}{Siedentopf} & \multicolumn{3}{c}{Stone} & \multicolumn{3}{c}{Straw}\\
                               &&  PSNR$\uparrow$ &SSIM$\uparrow$ & LPIPS$\downarrow$ & PSNR$\uparrow$ &SSIM$\uparrow$ & LPIPS$\downarrow$ & PSNR$\uparrow$ &SSIM$\uparrow$ & LPIPS$\downarrow$ & PSNR$\uparrow$ &SSIM$\uparrow$ & LPIPS$\downarrow$ & PSNR$\uparrow$ &SSIM$\uparrow$ & LPIPS$\downarrow$ & PSNR$\uparrow$ &SSIM$\uparrow$ & LPIPS$\downarrow$  \\
    \midrule
    \midrule
    3DGS \cite{kerbl20233d} & 1.9 / 61 & 20.47 & 0.824 & 0.205 & 23.19 & \metrictablethird{0.818} & 0.195 & \metrictablethird{22.84} & \metrictablefirst{0.842} & \metrictablefirst{0.144} & 18.82 & 0.803 & 0.237 & 28.36 & 0.935 & 0.108 & 22.84 & \metrictablethird{0.842} & 0.144 \\
    Mip-Splatting \cite{yu2024mip}& 1.61 / 73 & 21.24 & 0.837 & 0.214 & 23.75 & \metrictablesecond{0.824} & \metrictablethird{0.192} & \metrictablesecond{22.85} & \metrictablethird{0.793} & 0.189 & 20.45 & 0.834 & \metrictablethird{0.205} & 29.91 & 0.928 & 0.114 & 21.23 & \metrictablesecond{0.863} & \metrictablethird{0.129} \\
    GS-W \cite{zhang2024gaussian}& 1.02 / 104 & 22.22 & \metrictablesecond{0.902} & \metrictablesecond{0.134} & 24.10 & 0.758 & 0.240 & 20.56 & 0.679 & 0.299 & 19.62 & 0.844 & 0.228 & \metrictablesecond{32.49} & \metrictablesecond{0.962} & \metrictablethird{0.076} & 23.23 & 0.763 & 0.173 \\
    WildGaussians \cite{kulhanek2024wildgaussians}& 1.47 / 101 & 21.37 & 0.852 & 0.192 & 20.71 & 0.734 & 0.293 & 19.62 & 0.601 & 0.412 & 21.01 & 0.826 & 0.254 & 32.01 & \metrictablethird{0.959} & 0.077 & \metrictablethird{23.87} & 0.809 & 0.135 \\
    NexusSplats \cite{tang2024nexussplats} & 1.19 / 96 & 21.59 & 0.848 & 0.219 & 22.61 & 0.731 & 0.326 & 20.24 & 0.633 & 0.341 & 20.32 & 0.828 & 0.235 & 30.41 & 0.935 & 0.129 & 20.82 & 0.706 & 0.270 \\
    DroneSplat \cite{tang2025dronesplat}& 1.67 / 101 & \metrictablethird{22.42} & 0.850 & \metrictablethird{0.144} & \metrictablethird{24.17} & 0.801 & \metrictablefirst{0.123} & 22.49 & 0.752 & \metrictablesecond{0.155} & \metrictablesecond{23.42} & \metrictablesecond{0.866} & \metrictablefirst{0.096} & 30.97 & 0.946 & \metrictablefirst{0.054} & \metrictablesecond{24.17} & 0.830 & \metrictablefirst{0.107} \\
    RobustSplat \cite{fu2025robustsplat}& 0.37 / 104 & \metrictablesecond{22.89} & \metrictablethird{0.890} & 0.179 & \metrictablesecond{24.21} & 0.749 & 0.344 & 21.32 & 0.656 & 0.350 & \metrictablethird{22.97} & \metrictablethird{0.859} & 0.247 & \metrictablethird{32.091} & 0.948 & 0.136 & 23.195 & 0.752 & 0.260 \\
    RefineSplat (Ours)  & 0.45 / 107 & \metrictablefirst{23.24} & \metrictablefirst{0.911} & \metrictablefirst{0.123} & \metrictablefirst{25.73} & \metrictablefirst{0.847} & \metrictablesecond{0.182} & \metrictablefirst{23.65} & \metrictablesecond{0.804} & \metrictablethird{0.182} & \metrictablefirst{25.11} & \metrictablefirst{0.914} & \metrictablesecond{0.132} & \metrictablefirst{34.16} & \metrictablefirst{0.975} & \metrictablesecond{0.069} & \metrictablefirst{25.58} & \metrictablefirst{0.878} & \metrictablesecond{0.116} \\
    \bottomrule
    \end{tabular}}
    \caption{Quantitative result on the Ambiguous wild dataset. Performance is highlighted by color from \metrictablethird{third} \metrictablesecond{to} \metrictablefirst{first}.}
    \label{sub_tab:ambi_quan_2}
    \vspace{-0.1em}
\end{table*}
\begin{table*}[!t]
    \centering
    \footnotesize
    \setlength{\tabcolsep}{0.001pt}
    \resizebox{\linewidth}{!}{
    \begin{tabular}{lcccccccccccccccccccccc}
    \toprule
    \multirow{2}{*}{Method}& \multirow{2}{*}{\shortstack{GPU hrs / FPS}}& \multicolumn{3}{c}{Mountain} & \multicolumn{3}{c}{Fountain} & \multicolumn{3}{c}{Corner} & \multicolumn{3}{c}{Patio} & \multicolumn{3}{c}{Spot} & \multicolumn{3}{c}{Patio-High}  \\
    & & PSNR$\uparrow$ & SSIM$\uparrow$ & LPIPS$\downarrow$ & PSNR$\uparrow$ & SSIM$\uparrow$ & LPIPS$\downarrow$ & PSNR$\uparrow$ & SSIM$\uparrow$ & LPIPS$\downarrow$ & PSNR$\uparrow$ & SSIM$\uparrow$ & LPIPS$\downarrow$ & PSNR$\uparrow$ & SSIM$\uparrow$ & LPIPS$\downarrow$ & PSNR$\uparrow$ & SSIM$\uparrow$ & LPIPS$\downarrow$  \\
    \midrule
    \midrule 
    Ha-NeRF \cite{mildenhall2021nerf} & - / $<$1 & 18.64 & 0.485 & 0.499 & 16.71 & 0.393 & 0.569 & 19.23 & 0.684 & 0.367 & 16.82 & 0.543 & 0.393 & 17.85 & 0.460 & 0.599 & 16.67 & 0.463 & 0.505 \\
    3DGS \cite{kerbl20233d} & 0.35 / 116 & 19.40 & 0.638 & 0.213 & 19.96 & 0.659 & \metrictablethird{0.185} & 20.90 & 0.713 & 0.241 & 17.48 & 0.704 & 0.199 & 20.77 & 0.693 & 0.316 & 17.29 & 0.604 & 0.363 \\
    Mip-Splatting \cite{yu2024mip} & 0.18 / 82 & 19.23 & 0.581 & 0.273 & 19.74 & 0.624 & 0.252 & 21.44 & 0.767 & 0.161 & 16.96 & 0.688 & 0.188 & 21.14 & 0.757 & 0.220 & 19.12 & 0.640 & 0.303 \\
    Splatfacto-W \cite{xu2024splatfacto} & 1.03 / 43.1 & 20.70 & 0.661 & \metrictablethird{0.169} & 20.37 & 0.662 & 0.187 & 21.53 & 0.739 & 0.241 & 15.58 & 0.491 & 0.536 & 20.03 & 0.683 & 0.324 & 15.58 & 0.491 & 0.536 \\
    GS-W \cite{zhang2024gaussian} & 1.03 / 105 & 19.43 & 0.596 & 0.299 & 20.06 & \metrictablesecond{0.723} & 0.274 & 22.17 & 0.793 & 0.155 & 19.90 & 0.681 & 0.260 & 17.13 & 0.608 & 0.409 & 19.90 & 0.681 & 0.260 \\
    WildGaussian \cite{kulhanek2024wildgaussians} & 1.56 / 94 & 20.43 & 0.653 & 0.255 & 20.81 & 0.662 & 0.215 & 24.16 & 0.822 & \metrictablefirst{0.045} & 21.44 & 0.800 & 0.138 & 23.82 & 0.816 & 0.138 & 22.23 & 0.725 & 0.206 \\
    SpotLessSplats \cite{sabour2024spotlesssplats} & 1.87 / 81 & 19.84 & 0.580 & 0.294 & 20.19 & 0.612 & 0.258 & 24.03 & 0.795 & 0.258 & 21.55 & \metrictablefirst{0.838} & \metrictablefirst{0.065} & 23.52 & 0.756 & 0.185 & 20.31 & 0.664 & 0.259 \\
    DeSplat \cite{wang2024desplat} & 0.84 / 106 & 19.59 & \metrictablethird{0.710} & 0.170 & 20.27 & 0.680 & \metrictablesecond{0.170} & \metrictablethird{26.05} & \metrictablethird{0.880} & \metrictablesecond{0.090} & 20.89 & 0.810 & \metrictablethird{0.110} & \metrictablefirst{26.07} & \metrictablefirst{0.900} & \metrictablefirst{0.090} & \metrictablethird{22.59} & \metrictablesecond{0.840} & \metrictablefirst{0.120} \\
    HybridGS \cite{lin2024hybridgs} & 0.98 / 112 & \metrictablesecond{21.73} & 0.693 & 0.284 & \metrictablethird{21.11} & 0.674 & 0.252 & 25.03 & 0.847 & 0.151 & \metrictablesecond{21.98} & 0.812 & 0.169 & 24.33 & 0.794 & 0.196 & 21.77 & 0.741 & 0.211 \\
    DroneSplat \cite{tang2025dronesplat} & 1.35 / 101 & \metrictablethird{21.23} & 0.687 & \metrictablefirst{0.162} & \metrictablefirst{21.54} & \metrictablethird{0.705} & \metrictablefirst{0.168} & 24.77 & 0.823 & 0.106 & \metrictablethird{21.85} & 0.816 & \metrictablesecond{0.107} & 24.37 & 0.821 & \metrictablesecond{0.095} & 22.53 & 0.778 & 0.181 \\
    RobustSplat \cite{fu2025robustsplat} & 0.75 / 116 & 21.15 & \metrictablesecond{0.737} & 0.201 & 21.01 & 0.701 & 0.207 & 25.88 & 0.876 & 0.144 & 21.63 & \metrictablethird{0.827} & 0.139 & 24.85 & 0.808 & 0.164 & 21.84 & 0.757 & 0.192 \\
    T-3DGS \cite{markin2024t} & 1.72 / 103 & 20.62 & 0.703 & 0.223 & 20.83 & 0.681 & 0.218 & \metrictablesecond{26.14} & \metrictablesecond{0.890} & 0.114 & 20.96 & 0.819 & 0.154 & \metrictablethird{25.84} & \metrictablesecond{0.893} & \metrictablethird{0.127} & \metrictablefirst{22.76} & \metrictablethird{0.831} & \metrictablethird{0.170} \\
    RefineSplat (Ours) & 0.93 / 115 & \metrictablefirst{22.62} & \metrictablefirst{0.772} & \metrictablesecond{0.167} & \metrictablesecond{21.51} & \metrictablefirst{0.732} & 0.188 & \metrictablefirst{26.42} & \metrictablefirst{0.897} & \metrictablethird{0.104} & \metrictablefirst{22.08} & \metrictablesecond{0.831} & 0.126 & \metrictablesecond{25.87} & \metrictablethird{0.869} & 0.142 & \metrictablesecond{22.71} & \metrictablefirst{0.841} & \metrictablesecond{0.159} \\
    \bottomrule
    \end{tabular}}
    \vspace{-0.1em} \\
    \caption{Quantitative result on the NeRF On-the-go dataset. Performance is highlighted by color from \metrictablethird{third} \metrictablesecond{to} \metrictablefirst{first}.}
    \label{sub_tab:onthego_quan}
    \vspace{-0.1em}
\end{table*}

\begin{table*}[!t]
  \centering
  \scriptsize
  \setlength{\tabcolsep}{0.001pt} 
  \renewcommand{\arraystretch}{1.00}
  \resizebox{\linewidth}{!}{
  \begin{tabular*}{0.95\textwidth}{@{\extracolsep{\fill}} l c *{3}{ccc} @{}}
    \toprule
    \multirow{2}{*}{Method} & \multirow{2}{*}{\shortstack{GPU hrs / FPS}} & \multicolumn{3}{c}{Brandenburg Gate} & \multicolumn{3}{c}{Sacre Coeur} & \multicolumn{3}{c}{Trevi-Fountain} \\
    & & PSNR$\uparrow$ & SSIM$\uparrow$ & LPIPS$\downarrow$ & PSNR$\uparrow$ & SSIM$\uparrow$ & LPIPS$\downarrow$ & PSNR$\uparrow$ & SSIM$\uparrow$ & LPIPS$\downarrow$ \\
    \midrule
    \midrule
    Ha-NeRF \cite{mildenhall2021nerf}    & 452 / $<$1    
    & 24.04 & 0.877 & 0.139 
    & 20.02 & 0.801 & 0.171 
    & 20.18 & 0.690 & 0.222 \\
    3DGS \cite{kerbl20233d}         & 2.2 / 57       
    & 19.33 & 0.884 & 0.132
    & 17.70 & 0.845 & 0.176
    & 17.08 & 0.714 & 0.241 \\
    Mip-Splatting \cite{yu2024mip}    & 0.18 / 82    
    & 20.46 & 0.890 & 0.126
    & 18.14 & \metrictablesecond{0.879} & 0.168
    & 17.52 & 0.747 & 0.218 \\
    Splatfacto-W \cite{xu2024splatfacto}  & 1.05 / 40.2  
    & 26.82 & 0.911 & 0.126
    & 22.56 & \metrictablethird{0.876} & 0.157
    & 22.28 & 0.766 & 0.235 \\
    GS-W \cite{zhang2024gaussian}   &  1.2 / 51      
    & 23.51 & 0.897 & 0.166
    & 19.39 & 0.825 & 0.211
    & 20.06 & 0.723 & 0.271 \\
    WildGaussian \cite{kulhanek2024wildgaussians}& 11.5 / 86 
    & \metrictablethird{27.77} & \metrictablethird{0.927} & 0.133
    & 22.56 & 0.859 & 0.177
    & \metrictablethird{23.63} & 0.766 & 0.228 \\
    SWAG \cite{dahmani2024swag} & 0.8 / 15
    & 26.33 & \metrictablesecond{0.929} & 0.139
    & 21.16 & 0.860 & 0.185
    & 23.10 & \metrictablefirst{0.815} & \metrictablesecond{0.208} \\
    NexusSplats \cite{tang2024nexussplats} & 6.81 / 88
    & 27.12 & 0.924 & 0.143
    & \metrictablethird{22.61} & 0.857 & 0.180
    & \metrictablefirst{23.92} & 0.761 & 0.253 \\
    DroneSplat \cite{tang2025dronesplat}  & 1.8 / 97  
    & 19.88 & 0.871 & \metrictablefirst{0.112}
    & 16.85 & 0.808 & \metrictablethird{0.143}
    & 16.71 & 0.661 & 0.276 \\
    RobustSplat \cite{fu2025robustsplat}  & 1.01 / 103
    & 19.74 & 0.887 & 0.146
    & 17.87 & 0.847 & 0.178
    & 17.63 & 0.213 & 0.727 \\
    ForestSplats \cite{park2026forestsplats}  & 7.9 / 122
    & \metrictablefirst{28.13} & \metrictablefirst{0.935} & \metrictablethird{0.118}
    & \metrictablesecond{23.84} & \metrictablethird{0.876} & \metrictablefirst{0.123}
    & 23.11 & \metrictablesecond{0.802} & \metrictablethird{0.212} \\
    RefineSplat (Ours) & 2.13 / 109
    & \metrictablesecond{27.81} & \metrictablesecond{0.929} & \metrictablesecond{0.117}
    & \metrictablefirst{24.16} & \metrictablefirst{0.889} & \metrictablesecond{0.137}
    & \metrictablesecond{23.71} & \metrictablethird{0.792} & \metrictablefirst{0.204} \\
    \bottomrule
  \end{tabular*}}
  \caption{Quantitative result on the Photo Tourism dataset. Performance is highlighted by color from \metrictablethird{third} \metrictablesecond{to} \metrictablefirst{first}.}
  \label{sub_tab:photo_quan}
\end{table*}

\begin{table*}[!t]
  \centering
  \scriptsize
  \setlength{\tabcolsep}{0pt} 
  \renewcommand{\arraystretch}{1.05}
  \resizebox{\linewidth}{!}{
  \begin{tabular*}{0.95\textwidth}{@{\extracolsep{\fill}} l c *{3}{ccc} @{}}
    \toprule
    \multirow{2}{*}{Method} & \multirow{2}{*}{\shortstack{GPU hrs / FPS}} & \multicolumn{3}{c}{SimingShan}  & \multicolumn{3}{c}{Sculpture}  \\
    & & PSNR$\uparrow$ & SSIM$\uparrow$ & LPIPS$\downarrow$ & PSNR$\uparrow$ & SSIM$\uparrow$ & LPIPS$\downarrow$ \\
    \midrule
    3DGS \cite{kerbl20233d} & 1.1 / 59
     & 18.14 & 0.749 & 0.317
     & 16.11 & 0.396 & 0.286 \\
    Mip-Splatting \cite{yu2024mip} & 0.8 / 63 
     & 19.78 & 0.742 & 0.295
     & 17.22 & 0.404 & \metrictablesecond{0.249} \\
    Ha-NeRF \cite{mildenhall2021nerf} & 32 / $<$ 1 
     & 22.26 & \metrictablethird{0.767} & 0.298
     & 18.35 & 0.412 & 0.274 \\
    Splatfacto-W \cite{xu2024splatfacto} & 0.6 / 71
     & 22.13 & 0.715 & 0.268
     & 17.49 & \metrictablethird{0.438} & 0.289 \\
    GS-W \cite{zhang2024gaussian} & 0.6 / 104
     & 22.41 & 0.706 & 0.243
     & 17.34 & 0.414 & \metrictablethird{0.271} \\
    WildGaussian \cite{kulhanek2024wildgaussians} & 1.3 / 94
     & 22.63 & 0.707 & \metrictablethird{0.239} & 17.15 & 0.371 & 0.383 \\
    DroneSplat \cite{tang2025dronesplat} & 1.5 / 104
     & \metrictablethird{23.16} & \metrictablethird{0.759} & \metrictablefirst{0.152}
     & \metrictablethird{18.48} & \metrictablesecond{0.489} & \metrictablefirst{0.225} \\
    RobustSplat \cite{fu2025robustsplat} & 0.57 / 109
     & \metrictablesecond{23.20} & 0.703 & 0.476
     & \metrictablesecond{19.14} & 0.447 & 0.461 \\
    RefineSplat (Ours) & 0.72 / 111
     & \metrictablefirst{24.08} & \metrictablefirst{0.801} & \metrictablesecond{0.226}
     & \metrictablefirst{19.64} & \metrictablefirst{0.515} & 0.299 \\
    \bottomrule
  \end{tabular*}}
  \caption{Quantitative result on the Drone Imagery dataset. Performance is highlighted by color from \metrictablethird{third} \metrictablesecond{to} \metrictablefirst{first}.}
  \label{sub_tab:supple_Drone_quan}
\end{table*}

%% file: supple_fig/sup_tab_memory.tex
\begin{table*}[t]
    \centering
    \scriptsize
    \setlength{\tabcolsep}{0pt} 
    \renewcommand{\arraystretch}{1.4}
    \resizebox{\linewidth}{!}{
    \begin{tabular*}{0.95\textwidth}{@{\extracolsep{\fill}} l c *{3}{ccc} @{}}
    \toprule
    Method & Porter & Humanoid & Jockey & Lounge & Monster & Artwork & Benches & Bust & Statuette \\
    \midrule
    \midrule
    3DGS \cite{kerbl20233d} & 447.08 & 413.04 & 396.69 & 247.98 & 277.27 & 103.08 & 380.86 & 472.09 & 370.94 \\
    Mip-Splatting \cite{yu2024mip} & 521.56 & 213.16 & 412.77 & 203.82 & 194.96 & 54.24 & 432.05 & 465.87 & 357.62 \\
    DroneSplat \cite{tang2025dronesplat} & 389.84 & 110.01 & 288.72 & 114.13 & 79.24 & 22.12 & 269.96 & 294.58 & 266.31 \\
    RobustSplat \cite{fu2025robustsplat} & 347.45 & 51.32 & 214.27 & 86.68 & 64.39 & 62.03 & 266.12 & 226.19 & 211.20 \\
    RefineSplat (Ours) & 306.91 & 43.24 & 165.83 & 85.13 & 25.80 & 20.31 & 227.64 & 187.64 & 193.71 \\
    \bottomrule
    
    \end{tabular*}}
    \caption{Comparison of memory efficiency on the Ambiguous wild dataset.}
    \label{sub_tab:memory_amb1}
    \vspace{-0.1em}
\end{table*}

\begin{table*}[t]
    \centering
    \scriptsize
    \setlength{\tabcolsep}{0pt} 
    \renewcommand{\arraystretch}{1.1}
    \resizebox{\linewidth}{!}{
    \begin{tabular*}{0.95\textwidth}{@{\extracolsep{\fill}} l c *{3}{ccc} @{}}
    \toprule
    Method & Bookstore &\; Room & \; Rose & \; Turtle& \;Sidentopf &\; Wood & \; Stone & \;\;Straw &\;\; Jewel \\
    \midrule
    \midrule
    3DGS \cite{kerbl20233d} & 348.39 & \;243.30 & \;276.89 &\; 311.95 & 331.80 & 145.98 & 152.18 & 550.00 & 247.71 \\
    Mip-Splatting \cite{yu2024mip} & 327.32 & \;173.06 & \;325.42 & \;371.53 & 225.57 & 174.80 & 79.83 & 595.35 & 274.48 \\
    DroneSplat \cite{tang2025dronesplat} & 249.59 & \;107.08 & \;231.89 & \;245.38 & 108.92 & 117.45 & 91.45 & 394.02 & 176.21 \\
    RobustSplat \cite{fu2025robustsplat} & 295.95 &\; 104.01 & \;205.61 & \;220.56 & 84.70 & 108.19 & 31.54 & 377.40 & 166.67 \\
    RefineSplat (Ours) & 263.56 & \;97.70 & \;197.57 &\; 203.87 & 79.37 & 102.69 & 29.68 & 336.50 & 153.37 \\
    \bottomrule
    \end{tabular*}}
    \caption{Comparison of memory size on the Ambiguous wild dataset.}
    \label{sub_tab:memory_amb2}
    \vspace{-0.1em}
\end{table*}

\begin{table*}[t]
    \centering
    \scriptsize
    \setlength{\tabcolsep}{0pt} 
    \renewcommand{\arraystretch}{1.2}
    \resizebox{\linewidth}{!}{
    \begin{tabular*}{0.95\textwidth}{@{\extracolsep{\fill}} l c *{3}{ccc} @{}}
    \toprule
    Method & Mountain & Fountain & Corner & Patio & Spot & Patio-High \\
    \midrule
    \midrule
    3DGS \cite{kerbl20233d} & 447.08 & 413.04 & 396.69 & 247.98 & 277.27 & 103.08 \\
    Mip-Splatting \cite{yu2024mip} & 521.56 & 213.16 & 412.77 & 203.82 & 194.96 & 54.24 \\
    DroneSplat \cite{tang2025dronesplat} & 389.84 & 110.01 & 288.72 & 114.13 & 79.24 & 22.12 \\
    RobustSplat \cite{fu2025robustsplat} & 347.45 & 51.32 & 214.27 & 86.68 & 64.39 & 62.03 \\
    RefineSplat (Ours) & 306.91 & 46.85 & 218.63 & 86.61 & 25.80 & 20.31 \\
    \bottomrule
    \end{tabular*}}
    \caption{Comparison of memory efficiency on the NeRF On-the-go dataset.}
    \label{sub_tab:memory_onthego1}
    \vspace{-0.1em}
\end{table*}

\begin{table*}[!t]
    \centering
    \scriptsize
    \setlength{\tabcolsep}{12pt} 
    \renewcommand{\arraystretch}{1.2}
    \begin{tabular*}{0.95\textwidth}{@{\extracolsep{\fill}} l c *{3}{ccc} @{}}
    \toprule
    \multirow{2}{*}{Method} & \multicolumn{3}{c}{Photo Tourism} \\
    & Brandenburg Gate & Sacre Coeur & Trevi-Fountain \\
    \midrule
    \midrule
    WildGaussians \cite{kulhanek2024wildgaussians} & 60.50 & 69.35 & 151.39 \\
    DroneSplat \cite{tang2025dronesplat} & 54.11 & 77.12 & 127.16 \\ 
    RobustSplat \cite{fu2025robustsplat} & 82.29 & 82.15 & 183.13 \\    
    RefineSplat (Ours) & 37.51 & 31.26 & 108.84 \\    
    \bottomrule
    \end{tabular*}
    \caption{Comparison of memory efficiency on the Photo Tourism dataset.}
    \label{sub_tab:memory_photo_drone_1}
    \vspace{-0.1em}
\end{table*}

%% file: supple_fig/sub_tab_notation.tex
\begin{table*}[!t]
    \centering
    \resizebox{\linewidth}{!}{
    \begin{tabular}{clc}
    \toprule
      Notation &Shape  &    Definition \\
    \midrule
    \midrule
    $\mathcal{\mu}$ & $\mathbb{R}^{3}$ & Position of 3D Gausssians \\
    $\Sigma$ & $\mathbb{R}^{3 \times 3}$  & Covariance matrix of 3D Gaussians \\
    $S$ & $S_{k} \in \mathbb{R}^{3}$ & Scaling matrix of Gaussians \\ 
    $R$ & $R_{k} \in \text{SO(3)}$ & Rotation matrix of Gaussians \\ 
    $\alpha$ & $\mathbb{R}^{1}$ & Opacity of 3D Gaussians \\
    $c$ & $\mathbb{R}^{3}$ & View-dependent color of 3D Gaussians \\
    $W$ & $\mathbb{R}^{3 \times 3}$ & Viewing transformation matrix\\
    $J$ & $\mathbb{R}^{2 \times 3}$ & Jacobian of the affine approximation of the projective transformation\\
    $C$ & $\mathbb{R}^{3}$ & The obtained pixel value of the rendered image \\
    $\hat{I}$ & $\mathbb{R}^{3 \times H \times W}$ & Rendered image \\
    $I^{gt}$ & $\mathbb{R}^{3 \times H \times W}$ & Ground truth image \\ 
    $N_{sh}$ &  & Spherical harmonics (SH) coefficients \\ 
    \midrule
    $\mathcal{M}_{global}$ & $\mathbb{R}^{H \times W}$ & Entropy-aware global mask \\
    $\mathcal{M}_{local}$ & $\mathbb{R}^{H \times W}$ & Entropy-aware local mask \\
    $\mathcal{M}_{final}$ & $\mathbb{R}^{H \times W}$ & Entropy-aware adaptive mask \\
    $\mathcal{B}_{3 \times 3}$ &  & Box filter\\
    $\mathcal{D}_{\theta} $ &  & DINO feature extractor \\
    $\mathcal{T}_{ent}$ &  & Adaptive entropy threshold \\
    $\mathcal{T}_{statistics}$ &  & Dynamic Threshold for the transient mask of DroneSplat \\    
    $\mathcal{T}_{hard}$ &  & Fixed value Threshold \\
    $\mathcal{T}_{over}$ &  & Overlap Ratio \\
    $\psi_{\theta}$ &  & MLP Layer \\
    $\mathcal{H}_i$ &  & Entropy \\
    $\mathcal{E}()$ &  & Shannon Entropy function \\
    $\mathcal{E}_i$ &  & Confidence for instances \\
    $\psi_{\theta}$ &  & MLP Layer \\
    {\small{$(\mu_k^{*}, c_k^{*}, \alpha_k^{*}, \sum_k^{*})$}} & & The new attributes of merged Gaussians \\
    \bottomrule
    \end{tabular}}
    \caption{For clarity and to avoid confusion in notation, we provided notation table to easily understand our method.}
    \label{sub_tab:notation}
\end{table*}